\documentclass[11pt,letterpaper]{article}

\usepackage[utf8]{inputenc}
\usepackage[T1]{fontenc}
\usepackage{amsmath, amssymb, amsthm, bm}
\usepackage{geometry}
\usepackage{setspace}
\usepackage{tabularx}
\usepackage{booktabs} 
\usepackage{graphicx}
\usepackage{float}
\usepackage{natbib}   
\usepackage[colorlinks=true, linkcolor=blue, citecolor=blue, urlcolor=blue]{hyperref}
\usepackage{titlesec}
\usepackage{abstract}
\usepackage{microtype}
\usepackage{enumitem}

\titleformat{\section}{\normalfont\large\bfseries}{\thesection.}{0.5em}{}
\titleformat{\subsection}{\normalfont\normalsize\bfseries\itshape}{\thesubsection.}{0.5em}{}

\title{\textbf{The Organization of Inference: Information, Resource Constraints, and AI Production}}

\author{
Yukun Zhang\\
\small The Chinese University\\
\small of Hong Kong\\
\small Hong Kong, China\\
\scriptsize\texttt{215010026@link.cuhk.edu.cn}
\and
Kemu Xu\\
\small University of Edinburgh\\
\small Edinburgh, United Kingdom\\
\scriptsize\texttt{s2749200@ed.ac.uk}
\and
Yishen Chen\\
\small The Chinese University\\
\small of Hong Kong, Shenzhen\\
\small Shenzhen, China\\
\scriptsize\texttt{yishenchen@link.cuhk.edu.cn}
}

\date{}

\begin{document}

\maketitle

\begin{abstract}
\noindent
The economic value of inference depends on how capacity and task information are distributed across stages of AI production. We study these organizational margins using controlled workflow experiments on externally verified software-engineering tasks. In two matched resource panels, direct execution records the same success rate of 59.6 percent at logical-token ceilings of 12,000 and 24,000, while success under information-constrained planning rises from 36.2 to 51.2 percent. The planning disadvantage narrows by 15.0 percentage points (95 percent task-cluster bootstrap interval: 4.2 to 25.8). A strict read-only planning campaign varies whether the planner sees the task issue. At 12,000 tokens, issue access raises success by about 16 percentage points over issue-hidden planning. Compared with direct execution, task-informed planning is about 10 points lower at 12,000 tokens; at 24,000 tokens, it shows a 29.6-point advantage. In the resource panels, direct execution uses substantially less than either ceiling, while the planning workflow's binding rate falls from 46.2 to 0.8 percent and downstream execution accounts for 89.9 percent of the increase in total use. Scale determines the capacity available to a system; workflow and information structure shape the productive value realized from it.
\end{abstract}

\newpage

\section{Introduction}
\label{sec:introduction}

As inference becomes an increasingly important variable input into artificial-intelligence production, a central economic question is how much inference capacity a system should receive. Structured reasoning, repeated sampling, search, and verification can convert additional inference into better performance \citep{wang2023selfconsistency,yao2023tree,snell2024scaling}. Multi-stage agents also decide where to spend that capacity: on planning, repository inspection, execution, verification, or recovery. When these activities share a finite allowance, an intermediate stage has an opportunity cost. Planning may improve decomposition and coordination \citep{wang2023plansolve,shen2023hugginggpt,erdogan2025planact}, while consuming resources that could support downstream action. This paper studies that allocation problem through the net contribution of a fixed pre-execution planning stage.

Two organizational margins guide the analysis. The resource margin concerns how the net value of planning changes when its shared inference constraint is relaxed. The information margin concerns whether task-defining information reaches the component consuming planning inference. A planner may produce coherent guidance of little practical value if it cannot see the issue that execution must resolve. These margins connect the value-of-computation perspective in rational metareasoning \citep{russell1991metareasoning} to the organization of multi-stage AI production. Recent work studies when to plan and how to allocate sampling effort between planning and execution \citep{paglieri2025whenplan,cui2026dream}; our comparisons evaluate fixed workflow contracts under specified resource and information conditions.

We use issue-hidden planning as a deliberately sharp treatment of an organizational information friction. Planner--executor systems separate guidance from environment-facing action \citep{erdogan2025planact}, and security architectures such as CaMeL impose component-level information boundaries \citep{debenedetti2025camel}. These systems motivate attention to information flow. Evidence that models use long contexts unevenly further motivates examining which information reaches each stage \citep{liu2024lostmiddle}. The issue-visibility intervention isolates one concrete allocation decision: supplying task-defining information to the planner as well as to execution.

The experiments use repository-level software-engineering tasks from a frozen, screened pool of 40 SWE-bench Verified instances \citep{jimenez2024swebench,openai2024verified}. Each task supplies an issue, repository, and external verification environment. Agents work in isolated workspaces and produce patches evaluated by an independent Docker-based harness. Externally verified success means passing the frozen executable checks \citep{openai2026verifiedlimits}. The sample contains 35 Django tasks and five tasks from four other repositories.

Three workflow labels organize the comparisons. Direct execution ($T_G$) begins task-facing work without a separate planning stage. Information-constrained planning ($T_{GP}$) first examines repository context with the issue hidden. In the resource panels, this planner has broad diagnostic tools and an instruction to avoid editing; in the strict information campaign, its tool allowlist enforces read-only access. Task-informed planning ($T_{GPA}$) adds issue visibility under that same strict planning contract. The symbols $G$, $P$, and $A$ denote generation (execution), planning, and task awareness. Execution receives the issue whenever that stage is reached, together with preceding planning messages and tool observations. Planning and execution draw on a shared logical ledger of newly introduced text, with billed API usage recorded separately.

The empirical design combines two sources of evidence. The resource panels compare $T_G$ and $T_{GP}$ at ceilings of 12,000 and 24,000 logical tokens. Each contains the complete 40-task, two-backend, three-replicate grid, totaling 480 outcomes with randomized run order. The strict 12k information panel compares all three policies across 720 assignments; one of the 1,200 strict-campaign endpoints is missing, and all results hold under both completions of that outcome (Section~\ref{subsec:evidentiary_architecture}).

The principal resource result is a workflow-dependent response to the higher ceiling. Direct execution resolves 143 of 240 assignments in each panel, or 59.6 percent. Information-constrained planning rises from 87/240 (36.2 percent) to 123/240 (51.2 percent). Its effect relative to direct execution changes from $-23.3$ to $-8.3$ percentage points, yielding
\begin{equation}
\label{eq:intro_budget_moderation}
\widehat{\Delta}_B=\widehat{\tau}_P(24{,}000)-\widehat{\tau}_P(12{,}000)=0.150.
\end{equation}
The 95 percent task-cluster bootstrap interval is $[4.2,25.8]$ percentage points ($p=.008$). The high-ceiling planning effect remains negative. The planning workflow shows a marked change in resource pressure: its binding rate falls from 46.2 to 0.8 percent, and mean execution use rises from 5,207 to 6,531 logical tokens. Direct execution has substantially more unused capacity at both ceilings.

Issue visibility improves production within the strict planning contract. At 12k, complete-case success is 70/240 (29.2 percent) under information-constrained planning and 109/239 (45.6 percent) under task-informed planning. The information effect is about $+16$ percentage points and remains significant after Holm correction. Direct execution records 133/240 (55.4 percent); the task-informed-minus-direct estimate is about $-10$ percentage points, with 95 percent intervals extending from about $-20$ to near zero. At 24k, task-informed planning reaches 82.5 percent success, 29.6 points above direct execution. Section~\ref{subsec:results_boundary} presents the direct-execution comparisons.

The process evidence suggests a mechanism. At 12k, an issue-hidden planner consumes roughly half the token budget on a plan that cannot target the actual problem; 46 percent of runs hit the budget limit. Relaxing the ceiling to 24k removes most binding and narrows the planning penalty from $-23$ to $-8$ points, but direct execution---which never faces the displacement cost---gains nothing. Providing the issue appears to redirect planning toward productive coordination: success rises 16 points at 12k and, at 24k where capacity is no longer scarce, task-informed planning surpasses direct execution by 30 points. These results suggest that the return to planning depends on the task information it receives and the capacity left for execution.

The paper makes three contributions. First, the matched resource comparison shows that the observed return to a larger inference allowance depends on workflow organization. Second, the strict sensitivity analysis quantifies the value of task information within a maintained planning stage. Third, the economic framework brings these results together through coordination, misdirection, and the opportunity cost of displaced execution. These channels explain why capacity and information allocation belong in the same production problem.

The analysis connects research on inference-time computation \citep{brown2024large,snell2024scaling,wu2025inferencescaling}, agent planning and orchestration \citep{yao2023react,shen2023hugginggpt,erdogan2025planact}, and organizational information processing \citep{alchian1972production,garicano2000hierarchies,radner1993organization,vanzandt1999realtime,bolton1994firm,dessein2006adaptive}. Their shared allocation problem is that intermediate activities consume scarce processing capacity and contribute to output through downstream actions. In the present setting, logical tokens provide an operational measure of workflow resources. The broader economic lesson is that scale determines the capacity available to a system, while workflow and information structure shape the productive value realized from it.

Section~\ref{sec:literature_review} develops the literature connections, and Section~\ref{sec:conceptual_framework} presents the economic framework. Section~\ref{sec:experimental_design} describes the design and inference procedures. Sections~\ref{sec:main_results}--\ref{sec:robustness_validation} report results, process evidence, and robustness. Sections~\ref{sec:design_implications}--\ref{sec:conclusion} discuss economic implications, limitations, and conclusions. The Online Appendix supplies implementation details and additional analyses.

\begin{figure}[!t]
\centering
\includegraphics[width=\textwidth]{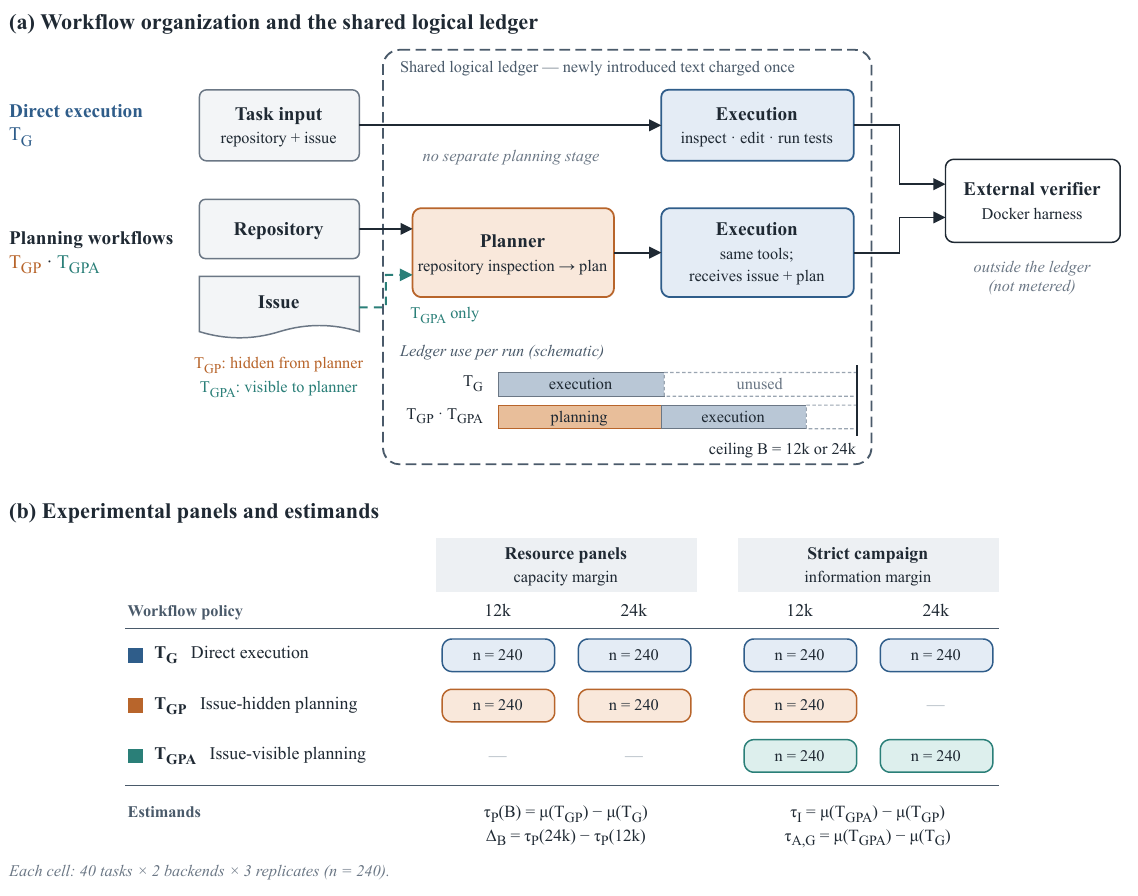}
\caption{Workflow organization and experimentally identified resource and information margins.}
\label{fig:main_concept}
\vspace{0.25em}
\begin{minipage}{\textwidth}
\footnotesize
\textit{Notes:} Panel (a) shows workflow inputs and the shared logical ledger. Execution uses the same tools across policies and receives the issue and any preceding planning context whenever that stage is reached. Planner tools are broad diagnostic in the resource panels and read-only in the strict campaign. Ledger-use bars are schematic; final artifact export and external verification follow the metered run. Panel (b) maps policies to experimental cells and estimands. Counts denote assignments; dashes mark unrun cells. Run order is randomized within each panel. The information contrasts $\tau_I$ and $\tau_{A,G}$ are defined at 12k.
\end{minipage}
\end{figure}

\section{Literature Review}
\label{sec:literature_review}

\subsection{Planning and Agentic Reasoning}
\label{subsec:literature_planning}

Structured intermediate representations can improve language-model performance
on tasks that resist a single-pass response. Chain-of-thought prompting elicits
stepwise rationales from worked demonstrations \citep{wei2022chain}; a minimal
zero-shot instruction can also elicit reasoning steps \citep{kojima2022zeroshot}.
Plan-and-Solve separates decomposition from execution
\citep{wang2023plansolve}, Least-to-Most solves an ordered sequence of simpler
subproblems \citep{zhou2023leasttomost}, and Decomposed Prompting delegates
subtasks to specialized modules \citep{khot2023decomposed}. Self-Ask generates
and answers follow-up questions, with optional external search
\citep{press2023compositionality}. These contributions document benchmark-specific gains from distinct intermediate representations, each with its own information and resource requirements.

Search methods explore several candidate reasoning states.
Tree of Thoughts evaluates alternative thoughts with lookahead and backtracking
\citep{yao2023tree}. RAP treats reasoning as planning with an internal world
model \citep{hao2023rap}, while Language Agent Tree Search combines tree search,
value estimates, reflection, and environmental feedback \citep{zhou2024lats}.
These methods show how structure can direct computation toward promising
paths. Their comparisons concern reasoning and search procedures; more recent
work also explicitly allocates computation across planning and execution.

Agent research grounds deliberation in environments where plans must lead to
actions. Zero-shot language-model planning maps high-level goals to admissible
actions \citep{huang2022zeroshotplanner}, ReAct interleaves reasoning with actions
and observations \citep{yao2023react}, and HuggingGPT uses an explicit
task-planning and tool-execution pipeline \citep{shen2023hugginggpt}.
Plan-and-Act separates a high-level planner from an environment-facing executor
for long-horizon tasks \citep{erdogan2025planact}. These systems clarify distinct
ways that intermediate guidance can coordinate tool-mediated actions; an
interleaved reasoning trace is not necessarily a separate pre-execution stage.

Feedback-based agents treat plans and outputs as revisable. Reflexion carries
linguistic feedback into later trials \citep{shinn2023reflexion}, whereas
Self-Refine uses model-generated critique to revise outputs
\citep{madaan2023selfrefine}. AdaPlanner and Inner Monologue revise behavior
using environmental feedback \citep{sun2023adaplanner,huang2023innermonologue}.
ProgPrompt constrains plans with programmatic state checks
\citep{singh2023progprompt}, and LLM-Planner replans when execution stalls or
fails \citep{song2023llmplanner}. DEPS uses failure feedback to correct plans
and a learned Selector to order subgoals by estimated completion steps
\citep{wang2023deps}. Voyager combines iterative environmental feedback with
an executable skill library \citep{wang2024voyager}. These mechanisms connect
intermediate guidance to the state encountered during execution.

Recent research directly examines planning costs. Learning When to Plan
compares fixed and dynamic planning policies and trains agents to decide when
to allocate computation to planning in long-horizon environments
\citep{paglieri2025whenplan}. SPIKE separates strategic planning from reactive
execution and uses event-triggered escalation to economize on expensive
reasoning \citep{jiang2026spike}. These studies establish planning frequency and
cost as substantive design choices.

\subsection{Inference-Time Computation under Resource Constraints}
\label{subsec:literature_compute}

Work on inference-time computation treats compute as an allocable input.
Self-consistency samples multiple reasoning paths and aggregates their answers
\citep{wang2023selfconsistency}; repeated sampling can increase the coverage of
correct candidate solutions \citep{brown2024large}. Adaptive-Consistency and
early-stopping self-consistency reduce sampling once the answer distribution
is sufficiently stable \citep{aggarwal2023adaptive,li2024esc}. These methods characterize allocation across repeated attempts. The return to
another sample depends on the problem, model, stopping rule, and ability to
select among candidates.

Search and verification provide other ways to direct test-time computation.
Process supervision trains models to assess intermediate mathematical steps
and supports candidate-solution selection \citep{lightman2024verify}.
Compute-optimal inference studies compare sampling, search, and verifier-guided
selection across models, problems, and budgets
\citep{snell2024scaling,wu2025inferencescaling}. In code generation,
feedback-guided search can improve search efficiency \citep{light2025sfs},
while PlanSearch explores natural-language plans to diversify candidate
programs \citep{wang2025plansearch}. Candidate coverage and successful selection
are distinct: finding at least one correct program does not ensure that the
returned program is correct. Nonzero verifier false-acceptance rates can limit
the gains from resampling \citep{stroebl2026limits}.

DREAM explicitly separates planning and execution search in mathematics and
code generation. Its reward-guided budget rule allocates sampling effort to
the two phases, stopping early on confident steps and spending more on
difficult ones \citep{cui2026dream}. PACE instead adapts reasoning-token budgets
to execution-time windows in embodied planning \citep{huang2026pace}. These frameworks connect allocation-rule design to the tradeoff between planning depth and execution capacity.

Agent studies increasingly measure operational costs as well as performance.
Agentic Plan Caching adapts stored plan templates to reduce the cost and
latency of repeated planning \citep{zhang2025plancaching}; LLMCompiler changes
orchestration to parallelize compatible function calls \citep{kim2024llmcompiler}.
FrugalGPT allocates queries across model APIs \citep{chen2024frugalgpt}, and
\emph{AI Agents That Matter} argues for evaluating accuracy jointly with cost
\citep{kapoor2025agents}. A cost- and latency-aware benchmark also compares
one-shot responses with tool-equipped plan--execute--replan agents, finding
task-dependent gains \citep{ghoshal2026toolsplanning}; that comparison changes
tools as well as planning and does not isolate planning alone. These approaches
optimize caching, orchestration, routing, or evaluation alongside task performance.

Rational metareasoning evaluates computational actions through their expected
effects on external decisions \citep{russell1991metareasoning}. It supplies the
conceptual basis for asking whether intermediate reasoning warrants its cost.
Organization economics adds a perspective on distributed processing.
Team-production theory studies joint output, metering, monitoring, and incentives
\citep{alchian1972production}; knowledge-hierarchy models study specialized
knowledge when communication and problem solving consume time
\citep{garicano2000hierarchies}. Models of decentralized information processing
make processing capacity and delay explicit
\citep{radner1993organization,vanzandt1999realtime}. Communication-network and
adaptive-organization models trade specialization or local adaptation against
communication and coordination \citep{bolton1994firm,dessein2006adaptive}.
Rational inattention provides another account of limited information-processing
capacity \citep{sims2003inattention}.

\subsection{Information Access and Externally Verified Agent Performance}
\label{subsec:literature_information}

Information-flow research motivates separating access from effective use.
CaMeL derives control from a trusted query while constraining flows from
untrusted tool outputs \citep{debenedetti2025camel}; Lost in
the Middle shows that the position of relevant context affects its use
\citep{liu2024lostmiddle}. These findings concern security boundaries and context placement rather than the allocation of task information to a separate planning component.

Interactive benchmarks evaluate agents through their consequences in an
environment. AgentBench spans environments requiring sequential decisions
\citep{liu2024agentbench}, WebArena evaluates tasks on functional websites
\citep{zhou2024webarena}, and OSWorld uses execution-based evaluators for
computer tasks \citep{xie2024osworld}. AppWorld checks application-state changes
with programmatic tests that permit multiple valid solutions
\citep{trivedi2024appworld}. Together, these benchmarks ground agent evaluation in environment-level consequences rather than intermediate outputs.

Code benchmarks sharpen this outcome-based perspective. HumanEval evaluates
generated functions with executable tests \citep{chen2021evaluating}, whereas
SWE-bench applies repository patches and requires task-associated tests to pass
\citep{jimenez2024swebench}. SWE-agent shows that the agent--computer interface
can change resolution rates \citep{yang2024sweagent}. Agentless provides a
complementary comparison through a fixed localization, repair, and validation
pipeline \citep{xia2025agentless}. Its simplicity highlights that workflow design choices alone can affect task resolution. SWE-Gym supplies executable repository tasks, agent trajectories,
and trained verifiers \citep{pan2025swegym}; SWE-Search examines tree search and
iterative refinement for issue resolution \citep{antoniades2025swesearch}.
These works connect repository evaluation to verifier-guided inference scaling
and search. Alongside PlanSearch, they motivate distinguishing candidate
generation, selection, and final test success when comparing compute use.

A well-formed plan is not itself the target outcome. SWE-bench operationalizes
success using task-associated FAIL\_TO\_PASS tests and PASS\_TO\_PASS regression
checks \citep{jimenez2024swebench}; AppWorld and OSWorld inspect resulting
application or computer state \citep{trivedi2024appworld,xie2024osworld}.
SWE-bench Verified adds human screening of the original benchmark instances
\citep{openai2024verified}. Subsequent analysis documents defective tests and
contamination risks \citep{openai2026verifiedlimits}. The \emph{externally verified success} endpoint means passing frozen
external checks, not complete software correctness or immunity to training-data
exposure.

Our experiments combine elements that appear separately across these literatures: fixed pre-execution contracts on repository tasks, a shared logical-token ceiling that creates competition between planning and execution, and a separate intervention on planner information access. The resource panels bring the value-of-computation perspective to complete agent workflows; the strict information panel measures the return to issue visibility within a maintained planning contract. Section~\ref{sec:experimental_design} describes the design and evidence status.

\section{Economic Framework}
\label{sec:conceptual_framework}

The framework organizes three forces: coordination benefits from planning, misdirection when guidance lacks task information, and the opportunity cost of execution capacity consumed by an intermediate stage. Appendix~\ref{app:continuous_planning} develops the corresponding continuous allocation model.

\subsection{Inference as an Organizational Resource}
\label{subsec:framework_allocable_input}

A conventional representation of AI production emphasizes model capability and aggregate inference capacity. Let $M$ denote model capability and $B$ the logical inference capacity assigned to a task. For a multi-stage agent, however, aggregate capacity does not fully describe the production environment. The system must also determine how inference is distributed across activities and what information is available to the components performing those activities.

We therefore use the reduced-form representation

\begin{equation}
\label{eq:organizational_production_framework}
Y
=
F(M,B,A,I,Z),
\end{equation}

where $Y$ denotes productive output, $A$ describes the organization of inference across stages, $I$ describes the allocation of task-relevant information across those stages, and $Z$ collects task, tool, repository, model-backend, and environmental characteristics.

Equation~\eqref{eq:organizational_production_framework} provides an organizing representation. The empirical design evaluates discrete workflow and resource contrasts within it; Appendix~\ref{app:continuous_planning} develops a theoretical extension to continuous allocation.

The relevant economic distinction is between the amount of inference available and the manner in which that inference is used. Two workflows may receive the same assigned capacity while placing different pressure on that capacity because they divide inference differently across intermediate and task-facing activities. Conversely, increasing the same nominal ceiling may have different productive consequences depending on whether it relaxes an active bottleneck inside the workflow.

\subsection{The Net Value of an Intermediate Stage}
\label{subsec:framework_opportunity_cost}

An explicit planning stage is an intermediate input rather than the final product. Its economic value is realized only through its effect on downstream execution. A useful decomposition is

\begin{equation}
\label{eq:planning_value_decomposition}
\begin{split}
\text{Net value of planning}
={}&
\text{coordination value}
-
\text{misdirection cost}
\\
&-
\text{opportunity cost of displaced execution}.
\end{split}
\end{equation}

The coordination component includes improvements in task decomposition, dependency identification, action ordering, repository prioritization, and downstream search. The misdirection component captures the possibility that a coherent plan may nevertheless direct execution toward irrelevant actions or parts of the environment. The opportunity-cost component arises because inference consumed before execution is not freely available for repository inspection, tool use, code modification, testing, or recovery from execution errors.

Direct execution can reason, adapt, and revise actions within task-facing execution. The planning treatment adds a separate stage that produces guidance before this work begins.

\medskip
\noindent
\textbf{Implication 1 (Opportunity cost of intermediate inference).}
\emph{An intermediate stage increases productive output only when the downstream coordination value it creates is sufficient to compensate for both planning-induced misdirection and the task-facing production capacity it consumes.}
\medskip

The implication yields no universal ranking between planning and direct execution. Planning may be productive in some environments and costly in others. Its net value depends on the information available to the planner, the scarcity of the common inference resource, the nature of the task, and the production activities displaced by the intermediate stage.

The empirical counterpart is the comparison between information-constrained planning and direct execution under a common assigned ceiling. This contrast measures the total production consequence of the evaluated policy, combining its coordination, misdirection, and resource-displacement channels.

\subsection{Information--Inference Matching}
\label{subsec:framework_information}

The productivity of intermediate inference depends on the informational environment of the stage that uses it. Processing capacity assigned to a planner may have limited value when the planner does not observe the information defining the problem that downstream execution must solve. The same planning architecture may become more productive when task-defining information is available to it.

In the strict information campaign, information-constrained planning and task-informed planning use the same read-only planning contract. Execution receives the task-specific issue when that stage is reached in either condition. The assigned difference is whether that issue is also visible during planning. This design isolates an information-allocation margin within the maintained planning architecture.

Task information may improve planning in at least two conceptually distinct ways. It may increase coordination value by allowing the planner to identify task-relevant files, dependencies, or actions. It may also reduce misdirection by preventing the planner from constructing guidance around irrelevant features of the repository.

\medskip
\noindent
\textbf{Implication 2 (Information--inference matching).}
\emph{Holding other conditions fixed, task information raises the net value of planning when its coordination benefits and reductions in misdirection outweigh any additional processing or displacement costs.}
\medskip

The issue-visibility comparison measures the gain from supplying task information to an existing planning stage. The planning-versus-direct comparison measures the net value of including that stage.

\subsection{Inference Scarcity and Workflow-Specific Returns}
\label{subsec:framework_scarcity}

The opportunity cost of an intermediate stage should also depend on the tightness of the common inference constraint. When inference capacity is scarce, planning and task-facing execution compete more sharply for the same resource. Capacity consumed before execution may then displace highly productive downstream activity. When the ceiling is relaxed, this displacement cost may become smaller.

Relaxing the inference constraint can attenuate a negative planning effect while leaving it below zero. This is the comparative prediction evaluated in the resource panels.

\medskip
\noindent
\textbf{Implication 3 (Inference scarcity and planning cost).}
\emph{When an intermediate stage places greater pressure on a common inference constraint, relaxing that constraint should attenuate the stage's opportunity cost relative to a workflow that already operates with greater resource slack.}
\medskip

The productive value of additional capacity depends on the internal constraint it relaxes. A higher ceiling may have substantial value for a workflow whose intermediate activities frequently encounter the resource limit, while producing little observed change in a workflow that usually terminates with unused capacity.

The experiment evaluates this implication using two discrete ceilings and fixed workflow policies. Its empirical counterpart is a workflow-dependent difference in the observed response to the higher ceiling.

\subsection{Empirical Mapping and Identification Boundaries}
\label{subsec:framework_empirical_mapping}

Table~\ref{tab:framework_mapping} maps the economic objects in the framework to the experimental contrasts.

\begin{table}[!htbp]
\centering
\small
\caption{Economic Objects and Experimental Contrasts}
\label{tab:framework_mapping}
\begin{tabular}{p{0.38\textwidth}p{0.23\textwidth}p{0.31\textwidth}}
\toprule
\textbf{Economic object}
&
\textbf{Contrast}
&
\textbf{Interpretation}
\\
\midrule

Net value of the evaluated information-constrained planning policy
&
$T_{GP}-T_G$
&
Effect of inserting the intermediate planning stage under a common inference ceiling
\\[0.7em]

Value of task-defining information within the maintained planning architecture
&
$T_{GPA}-T_{GP}$
&
Effect of making the task-specific issue available to the planner
\\[0.7em]

Moderation by inference scarcity
&
$\tau_P(24{,}000)-\tau_P(12{,}000)$
&
Change in the information-constrained planning effect when the assigned ceiling is relaxed
\\
\bottomrule
\end{tabular}

\vspace{0.4em}
\begin{minipage}{0.95\textwidth}
\footnotesize
\textit{Notes:}
$T_G$ denotes direct execution, $T_{GP}$ information-constrained planning,
and $T_{GPA}$ task-informed planning. The execution directive supplies the issue whenever execution is reached. The strict campaign shares a read-only contract between $T_{GP}$ and $T_{GPA}$; the resource panels use the broader diagnostic $T_{GP}$ contract.
\end{minipage}
\end{table}

The resource comparison evaluates how the net planning effect changes across two matched panels. The strict campaign evaluates issue visibility at the lower ceiling.

\section{Experimental Design and Evidence}
\label{sec:experimental_design}

The empirical program combines matched resource panels with a separate strict planning campaign. The resource panels compare two workflow policies at different logical-token ceilings. The strict campaign varies planner access to the task issue.

\subsection{Setting and Externally Verified Outcome}
\label{subsec:experimental_setting}

The experiments use a screened sample of 40 SWE-bench Verified tasks \citep{jimenez2024swebench,openai2024verified}. Screening ran direct execution three times per candidate at 12k using \texttt{deepseek-chat}. The frozen pool contained 160 candidates: 22 mid, 48 ceiling, and 90 floor tasks. The internal protocol targeted 40 non-floor tasks, restricting eligibility to candidates with at least one screening success. The configured selector prioritizes mid, low, high, then ceiling strata and retains the first 40 eligible identifiers in report order. With no low or high tasks, this selects all 22 mid tasks and the first 18 of the 48 ceiling tasks. All ceiling candidates scored 3/3, and their report order is lexicographic by task identifier.

The selected pool contains 35 Django tasks, two SymPy tasks, and one task each from Astropy, scikit-learn, and Sphinx. Treatment effects are averaged over this selected sample; Appendix~\ref{app:task_strata} describes the screening strata and Appendix~\ref{app:environment} provides the task-list reference.

Each task consists of an issue, a code repository, and an external verification environment. The agent works in an isolated workspace and submits a patch. For task $i$, recorded backend $m$, workflow $w$, and replicate $r$, the binary outcome is
\begin{equation}
\label{eq:verified_success_design}
Y_{imwr}=\mathbf{1}\{\text{the external SWE-bench harness marks the patch resolved}\}.
\end{equation}
The Docker-based harness evaluates the final patch independently of the agent's completion claim. A completed evaluation that is unresolved is a failure; an execution or verifier error that prevents a valid endpoint is recorded as missing.

\subsection{Workflow Treatments and Inference Capacity}
\label{subsec:experimental_treatments}

Table~\ref{tab:workflow_treatments} defines the workflow labels. The symbols retain the original implementation terminology: $G$ denotes generation, $P$ planning, and $A$ task awareness. We use ``execution'' for the editable generation stage, which can inspect files, modify code, and run tests.

\begin{table}[!htbp]
\centering\small
\caption{Workflow Policies and Campaign-Specific Planning Contracts}
\label{tab:workflow_treatments}
\begin{tabularx}{\textwidth}{@{}lXX@{}}
\toprule
\textbf{Policy} & \textbf{Workflow} & \textbf{Planner contract}\\
\midrule
$T_G$ & Direct execution & No separate planning stage\\[.4em]
$T_{GP}$ & Information-constrained planning, then execution & Issue hidden; broad diagnostic tools in resource panels; strict read-only tools in the information campaign\\[.4em]
$T_{GPA}$ & Task-informed planning, then execution & Issue visible; strict read-only tools\\
\bottomrule
\end{tabularx}
\par\smallskip
\begin{minipage}{\textwidth}\footnotesize
\textit{Notes:} The execution directive supplies the issue whenever execution is reached. The strict $T_{GP}$ and $T_{GPA}$ arms share the recorded planning-tool allowlist. The resource-panel $T_{GP}$ uses a different, broader tool contract.
\end{minipage}
\end{table}

In the resource panels, the planner receives an instruction to avoid editing and can use \texttt{list\_dir}, \texttt{read\_file}, \texttt{bash}, \texttt{run\_tests}, and \texttt{submit}. Direct editing tools are withheld, while shell access remains available. In the strict information campaign, the planning allowlist is restricted to \texttt{list\_dir}, \texttt{read\_file}, and \texttt{submit}. Its two planning arms vary issue visibility within that recorded contract. Planning messages and tool observations carry forward as context for execution. Appendix~\ref{app:workflow_contracts} gives the detailed contract and stage-reach table.

Planning and execution share one within-run logical ledger. Newly introduced stage directives, assistant outputs, and tool observations are charged once as they enter the workflow history. Assistant outputs use provider-reported completion-token counts; other text uses the harness counter, which attempts \texttt{cl100k\_base} tokenization and falls back to a character-based approximation. Repeated transmission of existing context is excluded from additional logical charges. Billed API input, output, and cache usage are recorded separately.

The ceilings are $B=12{,}000$ and $24{,}000$ logical tokens. The next model call's output limit is the smaller of its per-call stage cap and the remaining balance. The planning cap is 1,536 output tokens per call, with at most 12 turns per stage and earlier submission or budget stopping possible. The recorded binding flag is triggered by an exhausted/truncated budget or fewer than 32 tokens remaining. Appendix~\ref{app:logical_token_ledger} gives the accounting details. Final artifact export and external Docker evaluation follow the agent run; code editing and agent-side test observations occur inside it.

Execution is conditional on reaching that stage. Among 240 planning assignments in each arm, generation is reached in 233 resource-12k trials, all 240 resource-24k trials, 216 strict-12k $T_{GP}$ trials, and 173 strict-12k $T_{GPA}$ trials. These assignments remain in their assigned-policy analyses, including those that stop before execution.

\subsection{Experimental Panels and Estimands}
\label{subsec:experimental_estimands}

Each resource panel crosses 40 tasks, two API aliases, three replicates, and two policies ($T_G,T_{GP}$): 480 assignments, all with observed outcomes. The separately executed panels share the same task--backend--replicate design keys. The strict 12k panel includes all three policies (720 assignments, 719 observed endpoints); the strict 24k panel includes $T_G$ and $T_{GPA}$ (480/480).

Let $\mu_c(w)$ be mean verified success for workflow $w$ over campaign $c$'s frozen grid. For resource panels $R_{12},R_{24}$ and the strict information panel $S_{12}$, define
\begin{align}
\label{eq:main_estimands_design}
\tau_P(B)&=\mu_{R_B}(T_{GP})-\mu_{R_B}(T_G),\nonumber\\
\Delta_B&=\tau_P(24{,}000)-\tau_P(12{,}000),\nonumber\\
\tau_I&=\mu_{S_{12}}(T_{GPA})-\mu_{S_{12}}(T_{GP}),\nonumber\\
\tau_{A,G}&=\mu_{S_{12}}(T_{GPA})-\mu_{S_{12}}(T_G).
\end{align}
The resource comparison measures the net effect of the evaluated policy at each ceiling and its change across panels. The information contrast measures the value of issue visibility within the strict planning contract; $\tau_{A,G}$ compares that policy with direct execution. The strict protocol designates $\tau_{A,G}$ as primary and the information ($T_{GPA}-T_{GP}$) and planning-versus-direct ($T_{GP}-T_G$) contrasts as secondary.

\subsection{Protocol History and Evidence Status}
\label{subsec:evidentiary_architecture}

The internal resource protocol was frozen on June 30, 2026, and the strict protocol on July 14, 2026. The deviation plan was fixed on July 18 after bounded execution stopped and before arm-level effects were inspected. Table~\ref{tab:evidentiary_architecture} summarizes the evidence classifications.

\begin{table}[!htbp]
\centering\small
\caption{Experimental Panels and Evidence Status}
\label{tab:evidentiary_architecture}
\begin{tabularx}{\textwidth}{@{}XllX@{}}
\toprule
\textbf{Panel} & \textbf{Ceiling} & \textbf{Observed/assigned} & \textbf{Role}\\
\midrule
Protocol-frozen resource & 12k & 480/480 & Primary resource evidence\\
Protocol-frozen resource & 24k & 480/480 & Matched resource evidence\\
Strict information & 12k & 719/720 & Supporting; gate not met\\
Strict task-informed & 24k & 480/480 & Supporting; gate not met\\
\bottomrule
\end{tabularx}
\end{table}

The strict protocol requires valid outcomes for all 1,200 assignments and zero final errors. One $T_{GPA}$ assignment at 12k produced no valid endpoint: the agent requested a tool outside the planning allowlist, and retries did not resolve it. Because the completion requirement covers both ceilings jointly, neither strict panel qualifies as confirmatory. The deviation plan evaluates both binary completions of that missing outcome, $Y_{\mathrm{missing}}\in\{0,1\}$; conclusions must hold under both. Appendix~\ref{app:protocol_gates} documents the protocol and chronology; Appendix~\ref{app:endpoints} identifies the unit.

\subsection{Assignment Structure and Statistical Inference}
\label{subsec:statistical_inference}

Every included workflow covers the frozen task--backend--replicate grid. A seeded shuffle randomizes run order; every scheduled policy cell is executed. Both configurations use temperature 0.7 and three replicates.

Three inference methods are used. The resource protocol specifies a linear probability model with task and backend fixed effects and task-clustered standard errors as primary, with paired procedures as corroborating inference. The strict protocol specifies paired sign-flip as its primary test. Percentile task-cluster bootstraps provide confidence intervals. For each paired comparison, we average outcomes across backends and replicates within task, then take the mean of the 40 task differences.

At each strict endpoint, Holm adjustment applies to two secondary hypotheses: $T_{GPA}-T_{GP}$ and $T_{GP}-T_G$. The primary $T_{GPA}-T_G$ comparison additionally uses two one-sided $t$ tests (TOST) with an equivalence margin of $\pm10$ percentage points. Seeds, draw counts, and implementation details are in Appendix~\ref{app:additional_inference}; Appendix~\ref{app:holm} reports the Holm results; Appendix~\ref{app:equivalence_testing} reports the TOST results.

\subsection{Cross-Campaign Interpretation}
\label{subsec:cross_campaign}

The two resource ceilings were evaluated in separate matched panels; randomization determined run order within each panel. The contrast-of-contrasts $\Delta_B$ compares the workflow differences across ceilings, so a shift common to both workflows within a panel cancels.

Direct success changes little within each backend: 94/120 to 92/120 for \texttt{deepseek-chat}, and 49/120 to 51/120 for \texttt{glm-4.6} (Table~\ref{tab:app_model_heterogeneity}), while pooled planning success rises by 15.0 percentage points. A drift explanation would therefore need to improve planning disproportionately while leaving direct execution nearly unchanged in both backends. The process records also fit the budget-relief interpretation: planning-workflow binding falls from 46.2 to 0.8 percent, and downstream generation accounts for 89.9 percent of the increase in realized use (Section~\ref{sec:process_evidence}). These patterns support the resource interpretation. Because the ceiling was not randomized within a single panel, they do not exclude workflow-specific drift, but they narrow the alternative beyond general provider-side variation.

The recorded identifiers \texttt{deepseek-chat} and \texttt{glm-4.6} are hosted API aliases. Randomized run order distributes within-panel conditions across workflows; replication on frozen checkpoints would improve temporal reproducibility. The resource and strict information campaigns also use different planning-tool contracts. The information comparison is defined within the strict 12k campaign, while a common-contract information-by-ceiling interaction would require its missing $T_{GP}$ cell at 24k.

\section{Main Results}
\label{sec:main_results}

Results are reported as percentage-point differences in externally verified success. The resource panels provide the main evidence on inference capacity. The strict campaign estimates the effect of planner access to task information.

\subsection{Workflow-Dependent Returns to Relaxing Inference Scarcity}
\label{subsec:results_capacity}

Table~\ref{tab:main_budget_results} compares the two resource panels. Direct execution resolves 143/240 assignments at each ceiling. Information-constrained planning rises from 87/240 to 123/240. Consequently, the planning effect changes from $-23.3$ percentage points at 12k to $-8.3$ at 24k.

\begin{table}[!htbp] \centering \small \caption{Workflow Organization and the Relaxation of Inference Scarcity} \label{tab:main_budget_results}
\begin{tabular}{lccc} \toprule & \multicolumn{2}{c}{\textbf{Verified success}} & \\ \cmidrule(lr){2-3} \textbf{Workflow} & \textbf{12,000 tokens} & \textbf{24,000 tokens} & \textbf{Change} \\ \midrule Direct execution & $59.6\%$ & $59.6\%$ & $0.0$ pp \\[0.3em] Information-constrained planning & $36.2\%$ & $51.2\%$ & $+15.0$ pp \\ \midrule Planning effect & $-23.3$ pp & $-8.3$ pp & $+15.0$ pp \\ \bottomrule
\end{tabular} \vspace{0.4em}
\begin{minipage}{0.94\textwidth} \footnotesize \textit{Notes:} Each resource panel contains 480 observations, with 240 observations in each workflow arm. The planning effect is information-constrained planning minus direct execution. The final entry in the planning-effect row is the moderation estimand $\widehat{\Delta}_B = \widehat{\tau}_P(24{,}000) - \widehat{\tau}_P(12{,}000)$.
\end{minipage}
\end{table}

The moderation estimate is
\begin{equation}
\label{eq:budget_moderation_results}
\widehat{\Delta}_B=\widehat{\tau}_P(24{,}000)-\widehat{\tau}_P(12{,}000)=0.150.
\end{equation}
The fixed-effects estimate has a standard error of $0.057$ and $p=.008$. The 95\% task-cluster bootstrap interval is $[4.2,25.8]$ percentage points, and the task-level difference-in-differences sign-flip test gives $p=.013$.

The higher ceiling improves planning success while direct-execution success remains unchanged, narrowing the planning disadvantage by 15.0 percentage points.

\begin{figure}[!htbp] \centering \includegraphics[width=0.92\textwidth]{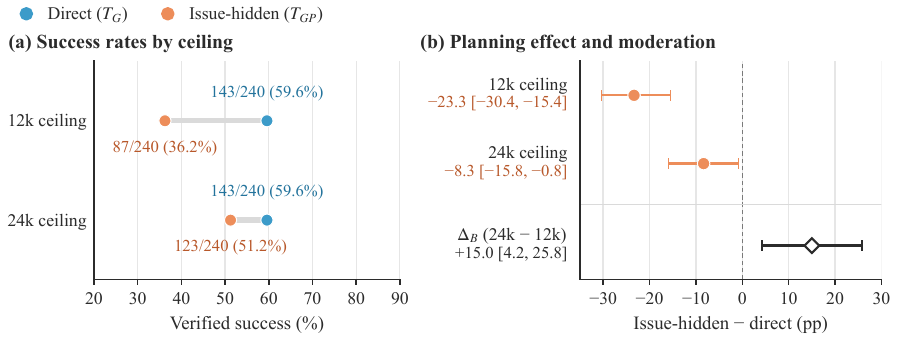} \caption{Workflow Organization and Inference Capacity} \label{fig:budget_moderation} \vspace{0.3em}
\begin{minipage}{0.92\textwidth} \footnotesize \textit{Notes:} Panel (a) shows observed success counts out of 240 per arm on a truncated rate axis; the gray segment at each ceiling spans the planning effect. Panel (b) shows $T_{GP}-T_G$ at each ceiling with 95\% task-cluster percentile-bootstrap intervals (2,000 resamples). Its bottom row shows the moderation estimate $\widehat{\Delta}_B=+15.0$ pp, with 95\% interval $[4.2,25.8]$ (20,000 task resamples) and task sign-flip $p=.013$. At 24k, the interval is below zero while sign-flip $p=.056$.
\end{minipage}
\end{figure}

Figure~\ref{fig:budget_moderation} displays descriptive arm rates and paired planning effects. The higher-ceiling estimate remains negative, with task sign-flip $p=.056$ and a 95\% task-bootstrap interval of $[-15.8,-0.8]$ percentage points. We report this individual contrast as sensitive to the inference procedure at the .05 threshold. The positive moderation estimate concerns the change in the planning effect across ceilings.

\subsection{The Low-Capacity Cost of Information-Constrained Planning}
\label{subsec:results_low_capacity}

At 12k, information-constrained planning reduces verified success by 23.3 percentage points relative to direct execution. The 95\% task-cluster bootstrap interval is $[-30.4,-15.4]$ percentage points, and the task sign-flip test yields $p=1.00\times10^{-5}$.

This estimate is the net workflow effect of inserting a resource-consuming stage with repository access, broad diagnostic tools, and the issue hidden. It combines any coordination benefits with misdirection, displaced execution, and other policy consequences. The negative estimate gives an empirical counterpart to the opportunity-cost framework in Section~\ref{subsec:framework_opportunity_cost}. Section~\ref{sec:process_evidence} examines the allocation of resources across stages.

\subsection{Matching Task Information with Planning Inference}
\label{subsec:results_information}

The strict 12k panel holds the planning-tool contract and assigned total ceiling fixed while varying issue visibility. Direct execution resolves 133/240 assignments (55.4\%); information-constrained planning resolves 70/240 (29.2\%); task-informed planning resolves 109/239 endpoints (45.6\%).

Table~\ref{tab:information_results_main} reports results under both completions of the missing outcome. The information effect is $+16.25$ to $+16.67$ percentage points ($p<.003$; 95\% bootstrap intervals above $+7$). Both conclusions survive Holm correction.

\begin{table}[!htbp] \centering \small \caption{Task Information and the Productivity of Planning} \label{tab:information_results_main}
\begin{tabularx}{\textwidth}{@{}p{.13\textwidth}p{.20\textwidth}p{.14\textwidth}p{.15\textwidth}X@{}} \toprule
\textbf{Missing value} & \textbf{Contrast} & \textbf{Effect (pp)} & \textbf{Sign-flip $p$} & \textbf{Bootstrap 95\% CI (pp)}\\
\midrule $Y_{\mathrm{missing}}=0$ & $T_{GPA}-T_{GP}$ & $+16.25$ pp & $0.00249$ & $[+7.1,+25.4]$ pp \\[0.4em] $Y_{\mathrm{missing}}=1$ & $T_{GPA}-T_{GP}$ & $+16.67$ pp & $0.00171$ & $[+7.9,+25.8]$ pp \\ \midrule $Y_{\mathrm{missing}}=0$ & $T_{GPA}-T_G$ & $-10.00$ pp & $0.09909$ & $[-20.4,0.0]$ pp \\[0.4em] $Y_{\mathrm{missing}}=1$ & $T_{GPA}-T_G$ & $-9.58$ pp & $0.11441$ & $[-20.0,+0.4]$ pp \\ \bottomrule
\end{tabularx} \vspace{0.4em}
\begin{minipage}{0.95\textwidth} \footnotesize \textit{Notes:} The strict information experiment is conducted under the 12,000-token logical inference ceiling. The first two rows compare task-informed planning with information-constrained planning. The final two rows compare task-informed planning with direct execution. The $T_{GPA}-T_{GP}$ conclusion remains significant under both completions and after Holm adjustment.
\end{minipage}
\end{table}

The protocol's other secondary contrast, information-constrained planning minus direct execution, is $-26.25$ percentage points under either completion (sign-flip $p\approx5\times10^{-6}$; bootstrap interval $[-33.33,-19.58]$). Appendix~\ref{app:holm} reports the full two-test family and its raw and adjusted values.

\begin{figure}[!htbp] \centering \includegraphics[width=0.92\textwidth]{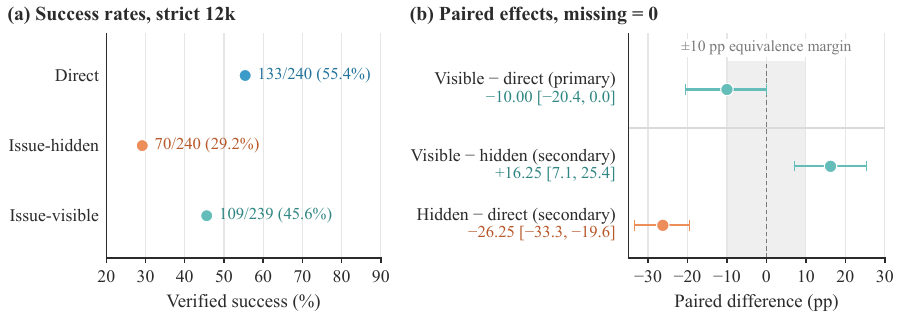} \caption{Matching Task Information with Planning Inference} \label{fig:task_information} \vspace{0.3em}
\begin{minipage}{0.92\textwidth} \footnotesize \textit{Notes:} Strict 12k campaign. Direct, issue-hidden, and issue-visible denote $T_G$, $T_{GP}$, and $T_{GPA}$. Panel (a) uses a truncated axis; the visible arm has 239 observed endpoints from 240 assignments. Panel (b) reports the three protocol contrasts, retaining all assignments with the missing endpoint set to failure ($Y_{\mathrm{missing}}=0$); bars are 95\% task-cluster percentile-bootstrap intervals (2,000 resamples), and the shaded band marks the $\pm10$ pp equivalence margin. Teal denotes contrasts involving issue-visible planning. Table~\ref{tab:information_results_main} and Figure~\ref{fig:appendix_protocol_sensitivity} report both completions.
\end{minipage}
\end{figure}

The information comparison measures the value of supplying the issue to the planning stage. Execution receives the issue in either planning arm; the treatment varies only whether planning also receives it. Figure~\ref{fig:task_information} shows the rates and contrasts.

\subsection{Task-Informed Planning versus Direct Execution}
\label{subsec:results_boundary}

Under the conservative completion ($Y_{\mathrm{missing}}=0$), the task-informed-minus-direct estimate is $-10.0$ percentage points (sign-flip $p=.099$; 95\% bootstrap interval $[-20.4,0.0]$). The other completion differs by less than 0.5~pp. The equivalence criterion ($\pm10$~pp) also fails (TOST $p=.50$). Table~\ref{tab:information_results_main} reports both completions. A substantial planning penalty remains possible, while a meaningful advantage is not supported.

At 24k, the strict panel observes all 480 outcomes. Task-informed planning resolves 198/240 assignments (82.5\%), compared with 127/240 (52.9\%) for direct execution, an advantage of 29.6 percentage points (95\% task-cluster bootstrap interval $[20.8,38.8]$). The sign reversal is consistent with the framework's resource interpretation: when capacity is scarce, a planning stage displaces execution; when capacity is sufficient and the planner is task-informed, planning can coordinate execution and raise output (Implication~3 in Section~\ref{subsec:framework_scarcity}; Appendix~\ref{app:optimal_planning_budget}). Because the 24k strict panel belongs to a protocol whose completeness gate failed at 12k, the result is retained as supporting evidence rather than a primary finding.

\section{Descriptive Process Evidence: Internal Inference Pressure}
\label{sec:process_evidence}

\begin{figure}[!t]
\centering
\includegraphics[width=0.92\textwidth]
{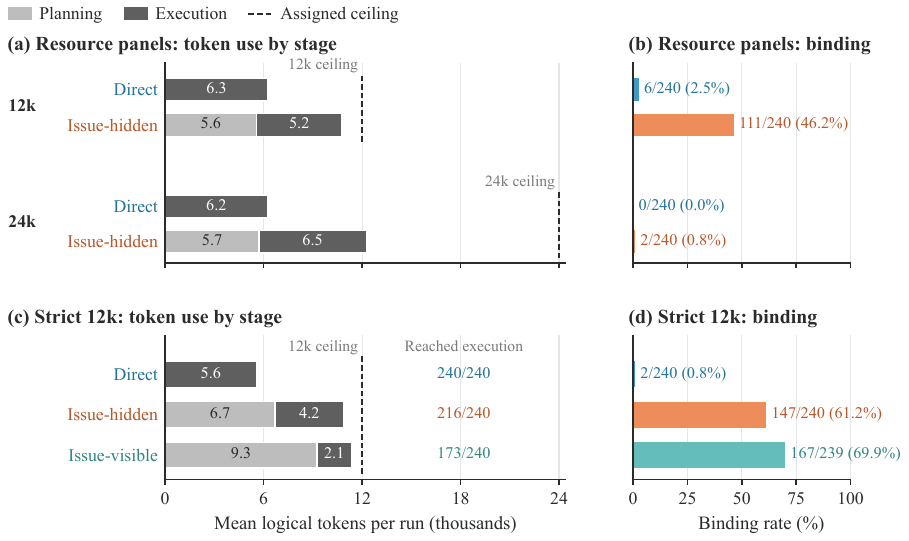}
\caption{Internal Inference Pressure and Realized Resource Use}
\label{fig:process_evidence}

\vspace{0.3em}
\begin{minipage}{0.93\textwidth}
\footnotesize
\textit{Notes:} Panels (a) and (c) show mean planning and execution-stage use per run in thousands of logical tokens (execution-stage use is recorded as generation in the ledger); gray shades denote stages and dashed lines mark the assigned ceiling. Panel (c) also lists the assignments reaching execution. Panels (b) and (d) use the same 0--100\% binding scale, with workflow colors as in Figure~\ref{fig:task_information}. Binding is the recorded exhaustion/truncation flag or a balance below 32 tokens. In the strict 12k campaign, issue-visible use and binding are computed over 239 observed runs. All panels are descriptive; campaign-specific planner contracts remain distinct.
\end{minipage}
\end{figure}

Realized resource use helps interpret the assigned-policy results. These are post-treatment measures: task difficulty, early stopping, tool activity, and success can all affect use. We therefore read them as descriptive process evidence, with the binding flag defined by the ledger's exhaustion or low-balance rule in Section~\ref{subsec:experimental_treatments}. Figure~\ref{fig:process_evidence} summarizes the patterns.

\subsection{Budget Binding}
\label{subsec:process_binding}

Under the resource experiment's 12k ceiling, binding occurs in 111/240 planning trials (46.2\%) and 6/240 direct-execution trials (2.5\%). At 24k these counts are 2/240 (0.8\%) and 0/240. The lower ceiling therefore places markedly greater pressure on the planning workflow.

The strict 12k panel shows a related pattern: binding occurs in 2/240 direct assignments (0.8\%), 147/240 information-constrained planning assignments (61.2\%), and 167/239 observed task-informed runs (69.9\%). Task-informed planning achieves higher success than issue-hidden planning while retaining frequent binding. Information access improves production within this resource-constrained workflow.

\subsection{Realized Inference Use and Downstream Generation}
\label{subsec:process_crowding}

In the resource panels, mean total use under information-constrained planning rises from 10,770 to 12,240 logical tokens. Mean planning use rises from 5,565 to 5,714, while execution-stage (``generation'' in the ledger) use rises from 5,207 to 6,531. Generation accounts for 89.9\% of the increase in total use.

Direct execution uses 6,267 tokens on average at 12k and 6,235 at 24k. Both means lie well below the assigned ceiling and accompany the same pooled success count. Within information-constrained planning, greater generation use accompanies a lower binding rate and a smaller performance disadvantage.

In the strict 12k panel, mean total use is 5,559 tokens for direct execution, 10,881 for issue-hidden planning, and 11,345 for task-informed planning. Planning use differs materially between the latter arms: 6,689 versus 9,256 tokens, with generation means of 4,192 and 2,089. The issue-visibility intervention changes the allocation across stages within a common ceiling: the task-informed planner spends more on planning, reaches execution less often, yet achieves higher overall success, suggesting that task information helps planning direct effort more productively.

Stage reach adds a descriptive view of this allocation. In the strict 12k panel, execution is reached in 173/240 issue-visible assignments and 216/240 issue-hidden assignments. Among those reaching execution, success is 109/173 (63.0\%) and 70/216 (32.4\%), respectively, compared with 133/240 (55.4\%) for direct execution. The 67 issue-visible assignments that stop earlier comprise 66 observed failures with a binding flag and the one missing endpoint. Conditioning on stage reach selects different realized subsets of assignments; the full-grid comparisons in Section~\ref{sec:main_results} retain all assignments.

In the resource panels, issue-hidden reach rises from 233/240 to 240/240, while success among reached assignments rises from 87/233 (37.3\%) to 123/240 (51.2\%). All seven unreached 12k assignments have observed unsuccessful outcomes. The improvement includes both more assignments reaching execution and a higher success rate among those that reach it.

\subsection{Resource Allocation across Workflows}
\label{subsec:process_interpretation}

Direct execution uses substantially less than either resource ceiling on average and records the same success count at both. The higher allowance therefore produces no observed gain in that workflow. The improvement occurs in the planning workflow, where binding falls from 46.2 to 0.8 percent and downstream generation accounts for 89.9 percent of the increase in realized use. This pattern supports an internal scarcity interpretation: the additional capacity is used mainly for task-facing execution in the workflow facing greater resource pressure.

The current comparisons combine displacement with changes in execution direction, context, tool use, and stopping time. To isolate the value of additional execution capacity, a further experiment could hold the completed plan and handoff context fixed and randomize the downstream execution allowance.

\section{Robustness and Validity}
\label{sec:robustness_validation}

The robustness exercises examine sensitivity to inference methods, the missing endpoint, and historical task samples. They preserve the campaign-specific evidence hierarchy in Table~\ref{tab:evidentiary_architecture}.

\subsection{Inferential Robustness}
\label{subsec:robustness_inference}

The moderation estimate is unchanged when the fixed-effects specification additionally absorbs task--model and replicate effects ($p=.010$ versus $.008$). All three resource estimands give the same directional conclusions under sign-flip, bootstrap, and fixed-effects inference. The individual 24k contrast is sensitive at the .05 threshold (sign-flip $p=.056$). Table~\ref{tab:app_resource_inference} in Appendix~\ref{app:resource_inference_summary} reports the methods side by side.

\subsection{Sharp Endpoints, Multiple Testing, and Equivalence}
\label{subsec:robustness_integrity}

Assigning the missing strict-12k outcome either zero or one preserves the positive information contrast and its Holm-adjusted significance. The task-informed-minus-direct estimate stays near $-10$ percentage points, and neither completion satisfies the equivalence criterion. Appendix~\ref{app:sharp_endpoint_inference} reports the endpoint calculations; Appendix~\ref{app:equivalence_testing} gives the two one-sided tests.

\subsection{Directional Stability across Task Samples}
\label{subsec:robustness_replication}

The negative planning effect also appears in experimental samples outside the main resource panel.

An earlier 12k planning-confirmation experiment, with 33 tasks and 392 clean outcomes from 396 assignments, produces an estimated planning effect of $-6.8$ percentage points. The fixed-effects estimate has a standard error of $0.023$ and $p=0.003$; the task-level sign-flip test gives $p=0.004$. Leave-one-task-out estimates range from $-7.6$ to $-6.0$ percentage points, preserving the negative sign after the removal of any single task.

A separately screened 12k independent pool of 23 tasks produces a larger negative estimate. The fixed-effects estimate is $-23.8$ percentage points, with a standard error of $0.061$ and $p<0.001$. The task-level estimate is $-24.8$ percentage points, and the task-cluster bootstrap $95\%$ confidence interval is $[-37.2,-14.5]$ percentage points.

Both historical samples use the same API aliases as the resource panels. The independent sample contains fewer than 30 task clusters and is classified as exploratory; it is not pooled with the main resource experiment.

The relevant conclusion is directional: across the main lower-ceiling panel, the planning-confirmation experiment, and the independent task pool, the planning effect remains negative. The sign is not an artifact of one task sample, though the magnitudes differ substantially ($-6.8$ versus $-23.3$ versus $-24.8$). The confirmation experiment uses a different task pool (33 SWE-bench tasks screened separately) and was run under an earlier evaluation-harness snapshot; either the task composition or the harness version could contribute to the smaller penalty. We do not pool across campaigns because of these design differences.

\subsection{Sample Integrity and Scope of the Robustness Evidence}
\label{subsec:robustness_scope}

Appendix~\ref{app:protocol_gates} records the coverage of each campaign. The two resource panels are complete (480/480 each); the strict campaign has one missing endpoint among 1,200 assignments. Historical samples are kept separate from the main resource comparisons.

\section{Economic Implications}
\label{sec:design_implications}

These results suggest that the return to a planning stage depends on the task information it receives and the capacity left for execution. Without the issue, the planner consumes scarce tokens on work that cannot target the actual problem; with the issue but insufficient capacity, displacement cost offsets coordination value; with both information and capacity, planning appears to coordinate execution and raise output. This section develops the economic implications of these conditions.

\subsection{Scale and the Organization of Inference}
\label{subsec:inference_organizational_resource}

\label{subsec:workflow_dependent_returns}
The resource panels show that the same nominal increase in capacity has different productive consequences across workflows. In organizational economics, intermediate activities consume scarce processing capacity and influence production through downstream actions \citep{alchian1972production,garicano2000hierarchies,radner1993organization,vanzandt1999realtime}. The inference-allocation analogue is direct: a planning stage that binds a shared token budget displaces execution, and relaxing that budget relieves the bottleneck only in the workflow where binding occurs. For agent evaluation, this suggests reporting resource ceilings alongside outcomes and realized use, so that comparisons across conditions can distinguish resource pressure from net workflow contribution.

\subsection{Information--Inference Matching}
\label{subsec:information_inference_matching}

The strict analysis shows the value of matching task information to the stage consuming planning inference. Its issue-visibility intervention improves success within a shared planning-tool contract and ceiling. Execution receives the issue whenever it is reached; the treatment determines whether planning also receives it.

Task information can guide planning toward relevant actions and away from misdirected work. A system can possess the information needed for a task while allocating it differently across stages; the intervention changes which component can use it.

Task information improves the planning workflow at 12k, while the task-informed-minus-direct contrast changes from about $-10$ points at 12k to $+29.6$ points at 24k. For agent design, the relevant question is both what information a planning stage receives and how much capacity remains for execution.

\subsection{From Fixed Workflows to Adaptive AI Production}
\label{subsec:adaptive_organization}

The fixed-contract results motivate a broader question: how should an agent adapt its workflow to task characteristics, available capacity, and information access? Direct execution may be attractive when coordination needs are limited; planning may be useful when decomposition has high value; verification and repair may become valuable after informative failures. These are candidate activities for an adaptive allocation rule.

Existing work on learned planning decisions and adaptive stage budgets addresses aspects of this problem \citep{paglieri2025whenplan,cui2026dream}. The present evidence contributes by showing how the observed performance of fixed contracts changes with their resource and information conditions. Stronger models and larger allowances can expand the feasible set of activities, while allocation rules determine how that set is used.

A natural next design would jointly vary information access, inference ceilings, and task characteristics within common campaigns using immutable model checkpoints. Evaluating workflow selection under that design would help connect the conditional effects reported here to empirically grounded rules for adaptive inference allocation.

\section{Discussion and Limitations}
\label{sec:limitations}

\paragraph{Domain and sample.}
The study evaluates repository-level software-engineering tasks with executable endpoints. Screening selects on pre-experimental behavior of the direct-execution control policy under one API alias. The pool contains 22 mid and 18 ceiling tasks, with 35 of 40 drawn from Django. The pooled effect weights these strata by their selected shares; Table~\ref{tab:app_task_strata} shows different descriptive effects across strata. The estimates therefore apply to this screened composition. Other domains, repositories, and task distributions may yield different effects.

\paragraph{Benchmark measurement.}
Externally verified success measures whether a patch passes the frozen harness. Subsequent analysis of SWE-bench Verified reports test defects and contamination risks \citep{openai2026verifiedlimits}. Its defect fraction concerns a selected set of frequently failed tasks; overlap with this study's pool and the hosted models' training exposure remain unverified. Test quality and differential use of prior exposure can affect the interpretation of measured workflow gains.

\paragraph{Workflow scope.}
The resource panels evaluate broad diagnostic planning under an instruction-level non-editing directive. The strict campaign evaluates issue visibility under a read-only allowlist. Interleaved planning, specialized planner models, and alternative search or repair architectures may have different resource requirements and production benefits. The estimates apply to the evaluated contracts and ceilings.

\paragraph{Resource measurement.}
The logical ledger combines provider-reported completion counts with a harness counter for other newly introduced text. It defines an operational constraint on workflow use. Physical computation, latency, and monetary expenditure require separate measurements.

\paragraph{Temporal and source reproducibility.}
The experiments record the API aliases \texttt{deepseek-chat} and \texttt{glm-4.6}. The resource comparison uses separately executed panels and assumes no provider-side changes that differentially affect the workflows (Section~\ref{subsec:cross_campaign}). The archived final records lack reliable absolute execution timestamps. In addition, the archived code for the strict task-informed workflow is incomplete: the tool contract and outcomes are fully documented, but not every runtime component is preserved. The archive supports offline statistical reproduction of all reported estimates; full runtime reconstruction would require additional materials. Appendix~\ref{app:workflow_contracts} documents this limitation.

\paragraph{Scope of identification.}
The design evaluates discrete workflow policies and two resource ceilings. The strict 24k panel lacks the issue-hidden $T_{GP}$ arm, so the design cannot identify an information-by-ceiling interaction under a common planning contract. The fixed workflow policies and two ceilings also leave optimal planning intensity and adaptive routing thresholds for future designs.

\section{Conclusion}
\label{sec:conclusion}

This paper studies how inference capacity and task information are organized across planning and execution. Doubling the token ceiling narrows the planning disadvantage by 15 percentage points while leaving direct execution unchanged; this additional capacity is used primarily for downstream execution. Issue visibility raises success by about 16 points within the strict planning contract, and at 24k, task-informed planning surpasses direct execution by 30 points.

The results suggest that the return to planning depends on the task information it receives and the capacity left for execution. Without the issue, planning appears to misdirect effort; without sufficient tokens, it displaces execution. These findings connect the productive value of inference to the workflow that uses it and the information available to each component. Scale determines the capacity available to a system; organization shapes the value realized from that capacity.

\paragraph{Data and Code Availability.}
The study archive contains trial-level outcomes, resource records, assignment grids, protocols, and analysis scripts sufficient for offline statistical reproduction. The strict task-informed runtime source is incomplete, as documented in Appendix~\ref{app:workflow_contracts}. The archive accompanies the study materials.

\clearpage
% --- 参考文献打印 ---
\bibliographystyle{aer}
\bibliography{paper2}

\clearpage
\appendix

% --- Appendix numbering ---
\renewcommand{\thesection}{\Alph{section}}
\renewcommand{\thesubsection}{\thesection.\arabic{subsection}}
\renewcommand{\thetable}{\thesection.\arabic{table}}
\renewcommand{\thefigure}{\thesection.\arabic{figure}}
\numberwithin{equation}{section}

\makeatletter
\@addtoreset{table}{section}
\@addtoreset{figure}{section}
\makeatother

\setcounter{section}{0}

\section*{Online Appendix}
\addcontentsline{toc}{section}{Online Appendix}

\noindent
This Online Appendix documents the experimental protocol, additional
statistical inference, sample integrity, extended economic framework,
descriptive heterogeneity, and supplementary figures underlying the results
in the main text. It introduces no new primary estimands. Its purpose is to
make the implementation and evidentiary hierarchy transparent and to
distinguish experimentally identified treatment effects from supporting,
exploratory, and theoretical extensions.

\medskip

\noindent
The appendix is organized as follows. Appendix~A describes the experimental
environment, workflow contracts, inference ledger, assignment grids, endpoint
construction, and campaign registry. Appendix~B reports additional statistical
inference. Appendix~C documents sample integrity and backend reproducibility.
Appendix~D develops the continuous allocation model omitted from the main text.
Appendix~E reports descriptive heterogeneity. Appendix~F collects supplementary
figures.

% ============================================================
% APPENDIX A
% ============================================================

\clearpage
\section{Experimental Protocol and Campaign Registry}
\label{app:experimental_protocol}

This appendix documents the experimental environment, assigned workflow
policies, logical inference ledger, assignment structure, endpoint definition,
and protocol hierarchy. The causal treatments are the assigned workflow and
assigned logical inference ceiling. Realized token use, budget exhaustion,
tool trajectories, and intermediate model behavior are post-treatment
quantities.

\subsection{Experimental Environment}
\label{app:environment}

The experiments use repository-level software-engineering tasks drawn from a
frozen pool of SWE-bench Verified instances
\citep{jimenez2024swebench}. Each task contains a natural-language issue, a
corresponding software repository, and an externally defined verification
environment.

Conditional on the assigned workflow, the agent operates in an isolated
workspace. It may inspect the repository, interact with the available tools,
modify files, and submit a final artifact. All workflow policies are evaluated
against the same task-specific external verification standard.

For task $i$, recorded model backend $m$, workflow $w$, and replicate $r$, the
primary outcome is

\begin{equation}
\label{eq:app_verified_success}
Y_{imwr}
=
\mathbf{1}
\left\{
\text{the submitted artifact passes external verification}
\right\}.
\end{equation}

Verification is performed through an independent Docker-based procedure rather
than by the agent itself. A model-reported declaration of completion, an
apparently coherent plan, or a syntactically plausible patch is insufficient
unless the resulting artifact satisfies the task-specific executable checks.

The outcome therefore measures realized production. Intermediate reasoning
receives credit only through its effect on the externally verified final
artifact. Table~\ref{tab:repository_composition} reports the selected pool's repository composition. The complete 40-task identifier list accompanies the study archive as \texttt{selected\_40\_task\_ids.txt}.

\begin{table}[!htbp]
\centering\small
\caption{Repository Composition of the Selected Task Pool}
\label{tab:repository_composition}
\begin{tabular}{lr}\toprule
\textbf{Repository} & \textbf{Tasks}\\\midrule
Django & 35\\
SymPy & 2\\
Astropy & 1\\
scikit-learn & 1\\
Sphinx & 1\\\midrule
Total & 40\\\bottomrule
\end{tabular}
\end{table}

\subsection{Workflow Contracts}
\label{app:workflow_contracts}

The planning contract is campaign-specific. The resource-panel planner receives an instruction to avoid editing and retains broad diagnostic tools, including \texttt{bash} and \texttt{run\_tests}. In the strict campaign, the recorded allowlist is \texttt{list\_dir/read\_file/submit}. The corresponding issue-hidden and issue-visible policies are denoted $T_{GP}$ and $T_{GPA}$.

\begin{table}[!htbp]\centering\small
\caption{Planning Contracts and Execution-Stage Reach}
\label{tab:app_workflow_contracts}
\begin{tabularx}{\textwidth}{@{}llXr@{}}\toprule
\textbf{Panel} & \textbf{Arm} & \textbf{Planning access} & \textbf{Execution reached}\\\midrule
Resource 12k & $T_{GP}$ & Issue hidden; diagnostic tools & 233/240\\
Resource 24k & $T_{GP}$ & Issue hidden; diagnostic tools & 240/240\\
Strict 12k & $T_{GP}$ & Issue hidden; strict allowlist & 216/240\\
Strict 12k & $T_{GPA}$ & Issue visible; strict allowlist & 173/240\\
Strict 24k & $T_{GPA}$ & Issue visible; strict allowlist & 240/240\\\bottomrule
\end{tabularx}
\par\smallskip
\begin{minipage}{\textwidth}\footnotesize
\textit{Notes:} Each corresponding $T_G$ arm reaches execution in 240/240 assignments. The 67 strict-12k $T_{GPA}$ assignments without execution include the single unresolved orchestrator error. Stage reach, recorded binding, and endpoint availability are distinct fields.
\end{minipage}
\end{table}

The resource planner records 1,817 \texttt{bash} and 45 \texttt{run\_tests} calls at 12k, and 1,824 and 48 at 24k. Its shell-command arguments are unavailable for an audit of workspace changes. The strict planning allowlists, issue-visibility flags, and directive hashes are directly recorded. Planning messages and tool observations pass to execution through the workflow history, and the execution directive introduces the full issue when that stage is reached.

The delivered strict snapshot lacks the complete $T_{GPA}$ runtime branch. Its contract comparison is supported by the protocol and recorded allowlists, visibility fields, and directive hashes; reconstructing every historical provider call would additionally require that runtime source. The archived outcomes and analysis scripts support offline statistical reproduction.

\subsection{Logical Inference Ledger}
\label{app:logical_token_ledger}

The assigned resource ceilings are
\begin{equation}\label{eq:app_budget_values}B\in\{12{,}000,24{,}000\}.
\end{equation}
The same \texttt{BudgetMeter} instance is shared across stages. Its non-overlapping charges are summarized below.
\begin{table}[!htbp]\centering\small
\caption{Logical-Ledger Accounting and Stopping Rules}
\label{tab:ledger_rules}
\begin{tabularx}{\textwidth}{@{}lX@{}}\toprule
\textbf{Component} & \textbf{Rule}\\\midrule
Assistant output & Provider-reported completion tokens, charged once.\\
New non-model text & Stage, system, and memory directives and tool observations use the harness \texttt{cl100k\_base} counter; on failure, $\lceil\mathrm{len(text)}/4\rceil$, at least one for nonempty text. Empty text counts zero.\\
Repeated context & Previously charged history is sent again without an additional logical charge.\\
Next output allowance & Minimum of remaining ledger balance and per-call stage cap; planning cap 1,536, stage turn limit 12.\\
Text over balance & Truncate to the available allowance and mark the meter exhausted.\\
Recorded binding & Meter exhausted, or remaining balance below 32 tokens.\\
Billed API usage & Provider input/output/cache counters recorded separately.\\
External evaluation & Final patch export and Docker scoring follow the metered agent run.\\\bottomrule
\end{tabularx}
\end{table}

Agent-side tests produce observations inside the execution stage; the final external evaluation scores the exported patch. The ledger's general role vocabulary also accommodates evaluation, repair, memory/management, and verification in other workflows. These role names describe accounting categories, and the active policies determine which stages run.

Assistant counts retain backend-specific tokenization, while non-model text uses the shared harness approximation. Historical records preserve charges and role totals but omit a per-message indicator of tokenizer versus fallback use. The measure therefore describes the operational text allowance implemented by this harness. Realized use can be below the assigned ceiling:
\begin{equation}\label{eq:app_assigned_realized}\text{realized logical use}\leq B.
\end{equation}
Per-call caps, early submission, the stage turn limit, and global ledger stopping jointly determine realized stage use.

\subsection{Assignment Grids and Estimands}
\label{app:randomization}

Let

\begin{equation}
c=(i,m,r)
\end{equation}

index a task, recorded model backend, and replicate. Let $Y_c(w,B)$ denote the
potential externally verified outcome under workflow $w$ and inference ceiling
$B$.

Workflow assignments are fixed before execution. They do not depend on
realized token use, intermediate model behavior, tool trajectories, or eventual
task success.

\paragraph{Protocol-frozen resource experiment.}

The resource experiment contains 40 tasks, two model
backends, three replicates, and two workflow policies, $T_G$ and $T_{GP}$. At
each inference ceiling, the assignment grid contains

\begin{equation}
40\times2\times3\times2
=
480
\end{equation}

trials. Both the $12{,}000$- and $24{,}000$-token panels contain all 480
externally observed outcomes.

For ceiling $B$, the planning estimand is

\begin{equation}
\label{eq:app_planning_estimand}
\tau_P(B)
=
\mathbb{E}_c
\left[
Y_c(T_{GP},B)
-
Y_c(T_G,B)
\right].
\end{equation}

The resource-moderation estimand is

\begin{equation}
\label{eq:app_budget_estimand}
\Delta_B
=
\tau_P(24{,}000)
-
\tau_P(12{,}000).
\end{equation}

A positive value of $\Delta_B$ means that the net effect of
information-constrained planning becomes less negative or more positive when
the common inference ceiling is relaxed. It does not require
information-constrained planning to outperform direct execution at the higher
ceiling.

\paragraph{Strict information experiment.}

The strict $12{,}000$-token information experiment uses the same 40 tasks, two
model backends, and three replicates, but includes all three workflow
policies. Its frozen assignment grid contains

\begin{equation}
40\times2\times3\times3
=
720
\end{equation}

trials. Of these, 719 have externally observable endpoints.

The secondary information estimand is

\begin{equation}
\label{eq:app_information_estimand}
\tau_I
=
\mathbb{E}_c
\left[
Y_c(T_{GPA},12{,}000)
-
Y_c(T_{GP},12{,}000)
\right].
\end{equation}

The paper additionally reports

\begin{equation}
\label{eq:app_aware_direct_estimand}
\tau_{A,G}
=
\mathbb{E}_c
\left[
Y_c(T_{GPA},12{,}000)
-
Y_c(T_G,12{,}000)
\right].
\end{equation}

The first contrast identifies the value of task information within the
maintained planning architecture. The second distinguishes that information
effect from the net value of maintaining a separate task-informed planning
stage relative to direct execution.

\subsection{Endpoint Definitions and Missing Outcomes}
\label{app:endpoints}

Binary success is the external SWE-bench harness resolved flag. Both resource panels contain 480 valid outcomes. The strict 12k panel has 719 valid outcomes from 720 assignments.

The missing assignment is at 12k, in arm $T_{GPA}$, replicate 0, using \texttt{deepseek-chat}. Its task is \texttt{django\_\_django-12858}; its trial identifier is \texttt{f24f597b619329d0}. Its final error is a prohibited planning-tool request for \texttt{grep}, rejected before workspace dispatch. Bounded retries of that intent remained unresolved. The record contains no valid external endpoint; complete-case analyses retain it as missing.

The pre-specified deviation plan evaluates both binary completions:
\begin{equation}\label{eq:app_sharp_endpoint}Y_{\mathrm{missing}}\in\{0,1\}.
\end{equation}
All 720 assignments enter each completed-outcome analysis. The width of the raw arm-mean effect across completions is exactly $1/240$, or 0.4167 percentage points. The same trial id also occurs at 24k with a valid outcome, because ids omit the budget field; endpoint identification therefore uses both id and ceiling. A protocol amendment corrects the identifier-matching rule to use both id and ceiling.

\subsection{Campaign Registry and Evidentiary Hierarchy}
\label{app:protocol_gates}

The resource protocol records a June 30, 2026 internal freeze. The strict protocol records July 14 and requires complete outcomes for 720 low-ceiling and 480 high-ceiling assignments, together with zero final errors across the combined campaign. Its final status is 719 low, 480 high, and one unresolved low-ceiling record. The July 18 deviation plan therefore classifies the combined campaign as incomplete under the strict confirmation requirements; the complete high-ceiling panel does not independently satisfy this joint requirement.

The July 18 deviation record states that the binary-completion procedure was fixed after bounded execution stopped and before arm-level results were inspected. These internal records supply the documented chronology; absolute execution dates cannot be reconstructed from the final trial schema. The registry below records completion and evidence status separately from the numerical contrasts.

\begin{table}[!htbp]\centering\small
\caption{Experimental Campaign Registry}
\label{tab:app_campaign_hierarchy}
\begin{tabularx}{\textwidth}{@{}XrrrX@{}}\toprule
\textbf{Campaign} & \textbf{Assigned} & \textbf{Observed/clean} & \textbf{Unavailable} & \textbf{Role}\\\midrule
Early five-arm & 330 & 195 & 135 & Exploratory\\
Planning confirmation & 396 & 392 & 4 & Supporting\\
Resource, 12k & 480 & 480 & 0 & Primary\\
Resource, 24k & 480 & 480 & 0 & Matched resource\\
Strict information, 12k & 720 & 719 & 1 & Supporting; gate not met\\
Strict task-informed, 24k & 480 & 480 & 0 & Supporting; gate not met\\\bottomrule
\end{tabularx}
\par\smallskip
\begin{minipage}{\textwidth}\footnotesize
\textit{Notes:} Unavailable denotes endpoints missing or ineligible for the historical clean analysis. The single strict-12k endpoint is unobserved, with zero assignments excluded from the two sharp-completion analyses. Earlier clean exclusions and this bounded missing endpoint have different analytical dispositions. The zero-final-error gate covers both strict panels jointly.
\end{minipage}
\end{table}

The resource contrast-of-contrasts combines the workflow differences at the two ceilings. Information effects use the strict-12k planning contract.

\section{Additional Statistical Inference}
\label{app:additional_inference}

This appendix reports the regression specifications and additional
task-structured inference underlying the estimates in the main text. The
procedures provide complementary uncertainty assessments. Causal identification
continues to come from assigned workflow and ceiling conditions rather than
from a particular regression specification.

\subsection{Fixed-Effects Specifications}
\label{app:fixed_effects}

For within-ceiling workflow comparisons, the complementary linear probability
specification is

\begin{equation}
\label{eq:app_workflow_fe}
Y_{imwr}
=
\alpha_i
+
\mu_m
+
\tau D^w_{imwr}
+
\varepsilon_{imwr},
\end{equation}

where $\alpha_i$ denotes task fixed effects, $\mu_m$ denotes recorded
model-backend fixed effects, and $D^w_{imwr}$ is the relevant assigned workflow
indicator. Standard errors are clustered at the task level.

For the resource-moderation analysis, the specification is

\begin{equation}
\label{eq:app_budget_interaction}
\begin{split}
Y_{imwrB}
={}&
\alpha_i
+
\mu_m
+
\rho High_B
+
\beta D^{GP}_{imwrB}
\\
&+
\delta
\left(
D^{GP}_{imwrB}
\times
High_B
\right)
+
\varepsilon_{imwrB},
\end{split}
\end{equation}

where $High_B=1$ under the $24{,}000$-token ceiling. The interaction coefficient
$\delta$ corresponds to the moderation estimand $\Delta_B$.

The baseline estimate is

\begin{equation}
\label{eq:app_budget_estimate}
\widehat{\Delta}_B
=
0.150,
\qquad
\mathrm{SE}=0.057,
\qquad
p=0.008.
\end{equation}

An alternative specification that additionally absorbs task--model and
replicate fixed effects yields the same point estimate,

\begin{equation}
\widehat{\Delta}_B
=
0.150,
\qquad
\mathrm{SE}=0.058,
\qquad
p=0.010.
\end{equation}

The estimated moderation effect is therefore not sensitive to the reported
fixed-effect specification.

\subsection{Task-Level Paired Inference}
\label{app:sign_flip}

The primary task-structured procedures begin with paired task-level treatment
contrasts. For the resource experiment, define

\begin{equation}
\label{eq:app_task_difference}
d_i(B)
=
\frac{1}{6}
\sum_{m=1}^{2}
\sum_{r=1}^{3}
\left[
Y_{im,T_{GP},r}(B)
-
Y_{im,T_G,r}(B)
\right].
\end{equation}

The task-level estimator is the average of $d_i(B)$ over the 40 tasks.

The two-sided sign-flip reference distribution assumes sign symmetry/exchangeability of task contrasts under the null. With 40 tasks the implementation uses 200,000 Monte Carlo draws (seed 12345), retaining zero differences. If $K$ draws are at least as extreme as the observed absolute mean, the reported value is $(K+1)/200001$. The minimum is approximately $5\times10^{-6}$; the resource-12k value near $10^{-5}$ corresponds to $K=1$. For comparisons with at most 22 task blocks the general implementation enumerates all sign assignments. This reference distribution is defined over task-difference signs; the experimental schedule randomizes run order.

For the moderation analysis, define

\begin{equation}
\label{eq:app_task_did}
q_i
=
d_i(24{,}000)
-
d_i(12{,}000).
\end{equation}

The paired sign-flip procedure evaluates the observed average task-level
contrast against the reference distribution obtained by reversing the signs of
the task-level differences while preserving their magnitudes. The procedure
treats the task as the inferential unit and preserves dependence among the two
model backends and three replicates within task.

For the lower-ceiling planning comparison,

\begin{equation}
\widehat{\tau}_P(12{,}000)
=
-0.233,
\qquad
p_{\mathrm{sign\text{-}flip}}
=
1.00\times10^{-5}.
\end{equation}

For the higher-ceiling comparison,

\begin{equation}
\widehat{\tau}_P(24{,}000)
=
-0.083,
\qquad
p_{\mathrm{sign\text{-}flip}}
=
0.056.
\end{equation}

For the change in planning effects,

\begin{equation}
\widehat{\Delta}_B
=
0.150,
\qquad
p_{\mathrm{sign\text{-}flip}}
=
0.013.
\end{equation}

The distinction between these tests is important. The principal resource
result concerns whether the planning effect changes across assigned ceilings.
It does not require the individual $24{,}000$-token planning contrast to be
statistically distinguishable from zero.

\subsection{Task-Cluster Bootstrap}
\label{app:bootstrap}

Bootstrap confidence intervals are constructed by resampling tasks with
replacement and retaining all workflow, recorded-backend, replicate, and
ceiling observations associated with each selected task. The procedure
therefore preserves within-task dependence.

Simple paired contrasts use 2,000 percentile task-cluster resamples: seed 12345 for resource-panel contrasts and 20260712 for strict endpoint contrasts. The task-level resource contrast-of-contrasts uses 20,000 resamples and seed 20260712. The 24k resource contrast has a 95\% interval of $[-0.158,-0.008]$. Its bootstrap interval and paired sign-flip $p=.056$ give different borderline decisions because they use different reference distributions.

For the $12{,}000$-token planning effect, the task-cluster bootstrap
$95\%$ confidence interval is

\begin{equation}
\label{eq:app_low_budget_bootstrap}
[-0.304,-0.154].
\end{equation}

For the moderation estimand, the reported 20,000-draw task-bootstrap
interval is

\begin{equation}
\label{eq:app_budget_bootstrap}
[0.042,0.258].
\end{equation}

The procedures agree on the lower-ceiling disadvantage and positive moderation. The individual 24k contrast is borderline across methods, as reported above.

\subsection{Summary of Protocol-frozen Resource Inference}
\label{app:resource_inference_summary}

Table~\ref{tab:app_resource_inference} places the complementary inference procedures side by side for the three resource estimands.

\begin{table}[!htbp]
\centering
\small
\caption{Protocol-frozen Resource Experiment: Additional Inference}
\label{tab:app_resource_inference}
\begin{tabularx}{\textwidth}{@{}p{.19\textwidth}p{.13\textwidth}p{.13\textwidth}p{.19\textwidth}X@{}}
\toprule
\textbf{Estimand} & \textbf{Effect (pp)} & \textbf{FE $p$} & \textbf{Sign-flip $p$} & \textbf{Task-bootstrap 95\% CI (pp)}\\
\midrule

$\tau_P(12{,}000)$
&
$-23.3$ pp
&
---
&
$1.00\times10^{-5}$
&
$[-30.4,-15.4]$ pp
\\[0.5em]

$\tau_P(24{,}000)$
&
$-8.3$ pp
&
---
&
$0.056$
&
$[-15.84,-0.83]$ pp
\\[0.5em]

$\Delta_B$
&
$+15.0$ pp
&
$0.008$
&
$0.013$
&
$[+4.2,+25.8]$ pp
\\

\bottomrule
\end{tabularx}

\vspace{0.4em}
\begin{minipage}{0.95\textwidth}
\footnotesize
\textit{Notes:}
The table reports only inferential quantities explicitly available from the
experimental analysis. An em dash indicates that the corresponding statistic
is not reported here. The fixed-effects moderation estimate has
$\mathrm{SE}=0.057$; the richer fixed-effect specification gives
$\mathrm{SE}=0.058$ and $p=0.010$.
\end{minipage}
\end{table}

\subsection{Multiple Testing and Holm Adjustment}
\label{app:holm}

The strict protocol specifies two secondary hypotheses at 12k: information visibility ($T_{GPA}-T_{GP}$) and planning versus direct ($T_{GP}-T_G$). Holm correction is applied to that two-hypothesis family at each endpoint. The primary $T_{GPA}-T_G$ comparison retains its own test. Table~\ref{tab:holm_results} reports the full secondary family.
\begin{table}[!htbp]\centering\small
\caption{Secondary Sign-Flip Tests and Holm Adjustment}
\label{tab:holm_results}
\begin{tabular}{@{}llrr@{}}\toprule
\textbf{Missing endpoint} & \textbf{Contrast} & \textbf{Raw $p$} & \textbf{Holm $p$}\\\midrule
0 & $T_{GPA}-T_{GP}$ & .00249499 & .00249499\\
0 & $T_{GP}-T_G$ & .00000500 & .00001000\\
1 & $T_{GPA}-T_{GP}$ & .00170999 & .00170999\\
1 & $T_{GP}-T_G$ & .00000500 & .00001000\\\bottomrule
\end{tabular}
\end{table}
The visibility test is second in the ordered two-test family, so its adjusted value equals its raw value. The $T_{GP}-T_G$ contrast is $-26.25$ percentage points under both completions, with 95\% task-bootstrap interval $[-33.33,-19.58]$. This is the other secondary contrast specified in the strict protocol.

\subsection{Sharp-Endpoint Calculations}
\label{app:sharp_endpoint_inference}

Table~\ref{tab:app_sharp_results} reports the strict information results under
both logically possible assignments of the single externally unobserved
task-informed outcome.

\begin{table}[!htbp]
\centering
\small
\caption{Sharp-Endpoint Sensitivity in the Strict Information Experiment}
\label{tab:app_sharp_results}
\begin{tabularx}{\textwidth}{@{}p{.13\textwidth}p{.20\textwidth}p{.14\textwidth}p{.15\textwidth}X@{}}
\toprule
\textbf{Missing value} & \textbf{Contrast} & \textbf{Effect (pp)} & \textbf{Sign-flip $p$} & \textbf{Bootstrap 95\% CI (pp)}\\
\midrule

$Y_{\mathrm{missing}}=0$
&
$T_{GPA}-T_{GP}$
&
$+16.25$ pp
&
$0.00249$
&
$[+7.1,+25.4]$ pp
\\[0.5em]

$Y_{\mathrm{missing}}=1$
&
$T_{GPA}-T_{GP}$
&
$+16.67$ pp
&
$0.00171$
&
$[+7.9,+25.8]$ pp
\\

\midrule

$Y_{\mathrm{missing}}=0$
&
$T_{GPA}-T_G$
&
$-10.00$ pp
&
$0.09909$
&
$[-20.4,0.0]$ pp
\\[0.5em]

$Y_{\mathrm{missing}}=1$
&
$T_{GPA}-T_G$
&
$-9.58$ pp
&
$0.11441$
&
$[-20.0,+0.4]$ pp
\\

\bottomrule
\end{tabularx}

\vspace{0.4em}
\begin{minipage}{0.96\textwidth}
\footnotesize
\textit{Notes:}
Source: the archived strict endpoint-sensitivity analysis under the July 18 deviation plan. Intervals use 2,000 task resamples with seed 20260712. The task-informed sharp rates are 109/240 (45.42\%) and 110/240 (45.83\%); the complete-case rate uses 109/239. The information contrast remains significant after Holm adjustment; Table~\ref{tab:holm_results} reports the full secondary family.
\end{minipage}
\end{table}

The sharp-endpoint analysis establishes

\begin{equation}
\tau_I>0
\end{equation}

under both $Y_{\mathrm{missing}}=0$ and $Y_{\mathrm{missing}}=1$.

By contrast, the comparison between task-informed planning and direct execution
remains unresolved under both assignments.

\subsection{Equivalence Testing}
\label{app:equivalence_testing}

The equivalence margin is $\pm0.10$ in success probability. The implementation forms 40 task-level paired differences, then conducts two one-sided $t$ tests with 39 degrees of freedom at $\alpha=.05$. The equivalence $p$-value is the larger of the two one-sided values. This decision corresponds to containment of the 90\% $t$ interval within the equivalence margins.

For $T_{GPA}-T_G$, the two sharp completions yield TOST $p=.500$ and $.471$, respectively. Neither meets the criterion. The effect estimates are $-10.00$ and $-9.58$ percentage points, and their 95\% task-bootstrap intervals permit a substantial disadvantage and at most a near-zero advantage. The margin is documented in the internal protocol; an external deployment-cost or power-based rationale is unavailable in the archived materials.

\subsection{Planning-Confirmation Stability}
\label{app:loto}

In the earlier planning-confirmation experiment, the estimated
information-constrained planning effect is $-6.8$ percentage points.

The task-and-recorded-model fixed-effects estimate has

\begin{equation}
\mathrm{SE}=0.023,
\qquad
p=0.003,
\end{equation}

while the paired task-level sign-flip procedure gives

\begin{equation}
p=0.004.
\end{equation}

Leave-one-task-out estimates range from

\begin{equation}
-7.6\text{ pp}
\quad\text{to}\quad
-6.0\text{ pp}.
\end{equation}

Every leave-one-task-out estimate preserves the negative sign. The analysis is
reported as supporting evidence on directional stability rather than as a new
primary estimand.

\subsection{Independent Task Pool}
\label{app:independent_pool}

A separately screened independent task pool contains 23 task clusters, of
which 22 provide complete task-level comparisons.

The task-and-recorded-model fixed-effects estimate is

\begin{equation}
\label{eq:app_independent_effect}
\widehat{\tau}^{\,\mathrm{ind}}_P
=
-0.238,
\end{equation}

with

\begin{equation}
\mathrm{SE}=0.061,
\qquad
p=9.15\times10^{-5}.
\end{equation}

The task-level estimate is

\begin{equation}
-0.248,
\qquad
p=3.66\times10^{-4},
\end{equation}

and the task-cluster bootstrap $95\%$ confidence interval is

\begin{equation}
[-0.372,-0.145].
\end{equation}

Because the analysis contains fewer than 30 task clusters, it remains
exploratory under the pre-specified guard. Its relevance is directional: the
evaluated information-constrained planning effect remains negative in a
separately screened task sample. The estimated magnitude is not pooled with
the resource result.

% ============================================================
% APPENDIX C
% ============================================================

\clearpage
\section{Sample Integrity and Reproducibility}
\label{app:sample_integrity}

Table~\ref{tab:app_campaign_hierarchy} is the campaign registry for assigned, observed/clean, and unavailable endpoints. Earlier clean exclusions differ from the single strict missing endpoint: the latter remains in both sharp-completion analyses. Both strict panels share the joint completeness gate described in Appendix~\ref{app:protocol_gates}.

\label{app:campaign_integrity}
\label{app:early_attrition}
\label{app:planning_confirmation_integrity}
\label{app:registered_completeness}
\label{app:high_budget_aware}
\label{app:backend_stability}
\label{app:cross_campaign_rule}

\paragraph{Primary panels.} Both resource panels are complete (480/480 each). The strict information campaign has one unobserved endpoint among 720 assignments; both sharp completions preserve the positive information effect. These completeness rates underlie the inferential hierarchy in Section~\ref{subsec:evidentiary_architecture}.

\paragraph{Supporting campaigns.} The early five-arm experiment has a $40.9\%$ drop rate (195 clean of 330 assigned), driven by orchestrator and API errors; it is retained only for exploratory analysis. The planning-confirmation experiment has a $1.0\%$ drop rate (392/396) and is used as supporting directional evidence but not pooled with the primary estimate.

\paragraph{Strict 24k panel.} All 480 outcomes are observed: 198/240 under $T_{GPA}$ (82.5\%) and 127/240 under $T_G$ (52.9\%), a difference of 29.6 percentage points (95\% task-cluster percentile-bootstrap interval $[20.8,38.8]$, 2,000 resamples). The panel belongs to the joint strict protocol whose completeness gate failed at 12k and is retained as supporting evidence.

\paragraph{Backend stability.} The experiments use \texttt{deepseek-chat} and \texttt{glm-4.6} API aliases. The archived trial schema lacks trustworthy absolute timestamps, so within-panel run-order randomization is the primary control for execution-time variation.

\paragraph{Cross-campaign interpretation.} Resource panels are separately executed matched grids whose workflow differences form $\Delta_B$. Within the strict 12k panel, issue visibility varies across a common planning allowlist. These two comparisons define the paper's resource and information margins.

\clearpage

\section{Additional Economic Framework}
\label{app:additional_framework}

This appendix develops a continuous planning-allocation problem that
motivates the opportunity-cost and scarcity interpretations. It is a theoretical extension of the framework in Section~\ref{sec:conceptual_framework}.

\subsection{Planning, Execution, and the Continuous Allocation Problem}
\label{app:continuous_planning}

Let $B>0$ denote total logical inference capacity and let $p\in[0,B]$ denote
the amount allocated to an explicit planning stage. Remaining task-facing
execution capacity is

\begin{equation}
\label{eq:app_execution_capacity}
e=B-p.
\end{equation}

Let $I$ denote the quality or availability of task-relevant information at the
planning stage, and let $Z$ collect task, model, tool, repository, and
environmental characteristics.

Productive value is represented as

\begin{equation}
\label{eq:app_productive_value}
\Pi(p;I,B,Z)
=
\Gamma(p,I;Z)
-
M(p,I;Z)
+
X(B-p;Z).
\end{equation}

The function $\Gamma$ represents coordination value generated by planning,
$M$ represents planning-induced misdirection, and $X$ represents productive
output from task-facing execution capacity. We normalize $\Gamma(0,I;Z)=M(0,I;Z)=0$, measuring coordination gains and misdirection costs relative to the absence of a separate planning stage.

We maintain the regularity conditions

\begin{equation}
\Gamma_p\geq0,
\qquad
\Gamma_{pp}\leq0,
\end{equation}

\begin{equation}
M_p\geq0,
\qquad
M_{pp}\geq0,
\end{equation}

and

\begin{equation}
X_e>0,
\qquad
X_{ee}\leq0.
\end{equation}

Planning may therefore generate coordination benefits with diminishing returns,
planning-induced misdirection may weakly increase with planning intensity, and
task-facing execution is productive with weakly diminishing marginal returns.

The marginal value of planning is

\begin{equation}
\label{eq:app_marginal_planning}
\Pi_p
=
\Gamma_p
-
M_p
-
X_e(B-p;Z).
\end{equation}

Planning raises productive value at the margin when

\begin{equation}
\label{eq:app_positive_planning}
\Gamma_p
-
M_p
>
X_e(B-p;Z).
\end{equation}

The condition makes the opportunity cost explicit: the marginal net
coordination return from planning must exceed the marginal productive value of
the execution capacity that planning displaces.

\subsection{Optimal Planning Intensity}
\label{app:optimal_planning}

A system designer that could continuously choose planning intensity would
solve

\begin{equation}
\label{eq:app_optimal_planning}
p^*(I,B,Z)
\in
\arg\max_{0\leq p\leq B}
\Pi(p;I,B,Z).
\end{equation}

For an interior solution, the first-order condition is

\begin{equation}
\label{eq:app_planning_foc}
\Gamma_p(p^*,I;Z)
-
M_p(p^*,I;Z)
=
X_e(B-p^*;Z).
\end{equation}

At the optimum, the marginal net coordination return from planning equals the
marginal value of inference remaining for task-facing execution.

The second derivative is

\begin{equation}
\label{eq:app_planning_second_derivative}
\Pi_{pp}
=
\Gamma_{pp}
-
M_{pp}
+
X_{ee}(B-p;Z).
\end{equation}

Under the maintained curvature assumptions,

\begin{equation}
\Pi_{pp}\leq0.
\end{equation}

A strictly negative value gives a locally unique interior optimum.

Direct execution, $p^*=0$, can therefore be optimal even when planning creates
positive coordination value. The relevant comparison is whether that value is
large enough to compensate for misdirection and displaced execution capacity.

\subsection{Information and Optimal Planning}
\label{app:optimal_planning_information}

Suppose task-relevant information raises the marginal coordination value of
planning,

\begin{equation}
\label{eq:app_gamma_information}
\Gamma_{pI}>0,
\end{equation}

and reduces the marginal cost of planning-induced misdirection,

\begin{equation}
\label{eq:app_m_information}
M_{pI}<0.
\end{equation}

Then

\begin{equation}
\label{eq:app_information_cross_partial}
\Pi_{pI}
=
\Gamma_{pI}
-
M_{pI}
>
0.
\end{equation}

Planning and task-relevant information are therefore complements in the
marginal production technology under these assumptions.

For an interior optimum satisfying $\Pi_{pp}<0$, the implicit-function theorem
gives

\begin{equation}
\label{eq:app_dp_di}
\frac{\partial p^*(I,B,Z)}{\partial I}
=
-
\frac{
\Pi_{pI}(p^*;I,B,Z)
}{
\Pi_{pp}(p^*;I,B,Z)
}
>
0.
\end{equation}

Better task information can therefore support a higher optimal planning
intensity in the continuous model.

The strict information panel varies issue visibility within a common
read-only planning contract and evaluates the total outcome contrast

\begin{equation}
\mathbb{E}
\left[
Y\mid T_{GPA},12{,}000
\right]
-
\mathbb{E}
\left[
Y\mid T_{GP},12{,}000
\right].
\end{equation}

This contrast is reported under both missing-outcome completions
in Section~\ref{subsec:results_information}. Equation~\eqref{eq:app_dp_di}
provides its theoretical allocation interpretation.

\subsection{Inference Capacity and Optimal Planning}
\label{app:optimal_planning_budget}

The cross-partial derivative of productive value with respect to planning
intensity and total inference capacity is

\begin{equation}
\label{eq:app_planning_budget_cross}
\Pi_{pB}
=
-
X_{ee}(B-p;Z).
\end{equation}

Under diminishing marginal returns to execution,

\begin{equation}
X_{ee}\leq0,
\end{equation}

and therefore

\begin{equation}
\Pi_{pB}\geq0.
\end{equation}

Inference capacity and planning intensity are thus complements in the marginal
objective under the maintained curvature assumptions.

For a locally unique interior optimum,

\begin{equation}
\label{eq:app_dp_db}
\frac{\partial p^*(I,B,Z)}{\partial B}
=
-
\frac{
\Pi_{pB}(p^*;I,B,Z)
}{
\Pi_{pp}(p^*;I,B,Z)
}
\geq0.
\end{equation}

Relaxing the total inference constraint can therefore support a larger optimal
planning allocation.

The resource panels compare a fixed information-constrained
planning policy at two discrete ceilings and estimate

\begin{equation}
\tau_P(24{,}000)
-
\tau_P(12{,}000),
\end{equation}

the change in the net planning effect as the resource ceiling increases.

\subsection{Discrete Net Planning Value}
\label{app:discrete_planning_value}

For a fixed positive planning policy, define the net value of planning relative
to direct execution as

\begin{equation}
\label{eq:app_net_planning_value}
\Delta_P(I,B,Z)
=
\Pi(p;I,B,Z)
-
\Pi(0;I,B,Z).
\end{equation}

Using Equation~\eqref{eq:app_productive_value},

\begin{equation}
\label{eq:app_net_planning_decomposition}
\Delta_P(I,B,Z)
=
\Gamma(p,I;Z)
-
M(p,I;Z)
+
X(B-p;Z)
-
X(B;Z).
\end{equation}

Differentiating with respect to total inference capacity gives

\begin{equation}
\label{eq:app_budget_moderation}
\frac{\partial\Delta_P(I,B,Z)}{\partial B}
=
X_e(B-p;Z)
-
X_e(B;Z)
\geq0,
\end{equation}

where the inequality follows from $B-p<B$ and $X_{ee}\leq0$.

For $B_H>B_L$,

\begin{equation}
\label{eq:app_discrete_budget_prediction}
\Delta_P(I,B_H,Z)
\geq
\Delta_P(I,B_L,Z).
\end{equation}

Equation~\eqref{eq:app_discrete_budget_prediction} permits a negative net
planning value at both ceilings: a larger ceiling can attenuate the penalty
while preserving its sign. This is the comparative implication evaluated by
the resource panels under the cross-panel condition above.

\subsection{Interpretive Scope}
\label{app:framework_scope}

The continuous model connects the two empirical margins to a broader
allocation problem. Its structural functions $\Gamma$, $M$, and $X$, optimal
planning choice $p^*$, and optimal-response derivatives remain theoretical
objects. The empirical estimates concern workflow outcome contrasts, their
change across separately executed resource panels, and issue visibility
within the strict planning contract. These comparisons support the
organization-of-inference interpretation developed in Section~\ref{sec:design_implications}.

The observed design covers two resource ceilings for $T_G/T_{GP}$ and all
three workflows at the strict low ceiling. Estimating a general
information-by-capacity response surface would require additional treatment
cells under a common contract.

% ============================================================
% APPENDIX E
% ============================================================

\clearpage
\section{Descriptive Heterogeneity}
\label{app:heterogeneity}

The heterogeneity analyses are descriptive extensions of the
average treatment effects. Subgroup estimates are not interpreted as
statistically established differences unless supported by a direct interaction
test.

\subsection{Recorded Model Backends}
\label{app:model_heterogeneity}

The resource experiment contains two API backends,
\texttt{deepseek-chat} and \texttt{glm-4.6}.
Table~\ref{tab:app_model_heterogeneity} reports the
information-constrained planning contrast separately by backend.

\begin{table}[!htbp]
\centering
\small
\caption{Planning Effects by Recorded Model Backend}
\label{tab:app_model_heterogeneity}
\begin{tabular}{llccc}
\toprule
\textbf{Ceiling}
&
\textbf{Recorded backend}
&
\textbf{Direct}
&
\textbf{Information-constrained}
&
\textbf{Difference}
\\
\midrule

12,000
&
\texttt{deepseek-chat}
&
$78.3\%$
&
$43.3\%$
&
$-35.0$ pp
\\[0.4em]

12,000
&
\texttt{glm-4.6}
&
$40.8\%$
&
$29.2\%$
&
$-11.7$ pp
\\

\midrule

24,000
&
\texttt{deepseek-chat}
&
$76.7\%$
&
$62.5\%$
&
$-14.2$ pp
\\[0.4em]

24,000
&
\texttt{glm-4.6}
&
$42.5\%$
&
$40.0\%$
&
$-2.5$ pp
\\

\bottomrule
\end{tabular}

\vspace{0.4em}
\begin{minipage}{0.95\textwidth}
\footnotesize
\textit{Notes:}
Backend names are runtime API aliases rather than guaranteed immutable
model-weight snapshots. The table reports descriptive subgroup contrasts.
\end{minipage}
\end{table}

For \texttt{deepseek-chat}, the planning contrast changes from $-35.0$
percentage points at $12{,}000$ tokens to $-14.2$ percentage points at
$24{,}000$ tokens.

For \texttt{glm-4.6}, the corresponding contrast changes from $-11.7$ to
$-2.5$ percentage points.

The higher ceiling therefore attenuates the negative planning contrast for both
backends. These subgroup patterns do not establish structural model
heterogeneity.

In the separate planning-confirmation experiment, the workflow-by-model
interaction estimate is

\begin{equation}
0.067,
\qquad
\mathrm{SE}=0.040,
\qquad
p=0.096.
\end{equation}

The available evidence is therefore insufficient to establish statistically
confirmed differences in treatment response across backends.

\subsection{Task-Screening Strata}
\label{app:task_strata}

Before the experiment, tasks were assigned to screening strata using
baseline direct-generation performance. The sample contains tasks
classified as \emph{mid} and \emph{ceiling}. These labels arise from the
screening procedure and should not be interpreted as cardinal measures of task
difficulty.

\begin{table}[!htbp]
\centering
\small
\caption{Planning Effects by Task-Screening Stratum}
\label{tab:app_task_strata}
\begin{tabular}{llcc}
\toprule
\textbf{Ceiling}
&
\textbf{Screening stratum}
&
\textbf{Information-constrained $-$ direct}
&
\textbf{Observations}
\\
\midrule

12,000
&
Ceiling
&
$-33.3$ pp
&
216
\\[0.4em]

12,000
&
Mid
&
$-15.2$ pp
&
264
\\

\midrule

24,000
&
Ceiling
&
$-5.6$ pp
&
216
\\[0.4em]

24,000
&
Mid
&
$-10.6$ pp
&
264
\\

\bottomrule
\end{tabular}

\vspace{0.4em}
\begin{minipage}{0.95\textwidth}
\footnotesize
\textit{Notes:}
Screening strata are based on the pre-experimental task-screening procedure.
The subgroup estimates are descriptive and are not interpreted as validated
treatment-effect heterogeneity.
\end{minipage}
\end{table}

At the lower ceiling, the negative planning contrast is larger in the ceiling
stratum. At the higher ceiling, the ceiling-stratum contrast is closer to zero,
while the mid-stratum contrast remains negative.

The pattern does not support a simple monotone relationship between screening
status and the return to planning. The strata are therefore reported
descriptively rather than as evidence of a validated task-difficulty threshold.

\subsection{Task-Level Treatment Differences}
\label{app:task_level_heterogeneity}

Task-level treatment differences vary around the average effects reported in
the main text. Figure~\ref{fig:appendix_task_differences} displays the
task-level information-constrained planning contrast under the
$12{,}000$- and $24{,}000$-token ceilings.

The figure is intended to show dispersion and sign variation rather than to
identify a latent task-level treatment rule. Each task-level difference is
estimated from a finite number of recorded-backend and replicate observations
and consequently contains sampling variation.

The paper therefore does not rank tasks by their realized effects or interpret
individual task estimates as stable treatment parameters. The identified
average result is that the planning contrast becomes materially less negative
when the common inference constraint is relaxed.

% ============================================================
% APPENDIX F
% ============================================================

\clearpage
\section{Additional Figures and Tables}
\label{app:additional_figures}

This appendix collects supplementary diagnostics supporting the protocol,
inference, heterogeneity, and sample-integrity analyses. The figures retain the
inferential status assigned in the main text and preceding appendices. They
should not be interpreted as additional primary treatment tests.

\par\medskip
\noindent\begin{minipage}{\textwidth}
\subsection{Sample Construction}
\label{app:fig_sample_construction}

Figure~\ref{fig:appendix_sample_construction} summarizes assigned and observed
endpoints across the experimental campaigns.

\begin{figure}[H]
\singlespacing
\centering
\includegraphics[width=0.94\textwidth]
{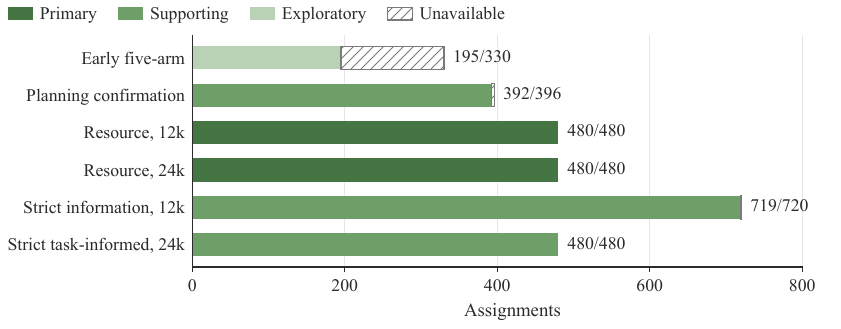}
\caption{Sample Construction and Endpoint Integrity}
\label{fig:appendix_sample_construction}

\vspace{0.3em}
\begin{minipage}{0.94\textwidth}
\footnotesize
\textit{Notes:} Counts reconcile assignments and valid/clean endpoints, as in Table~\ref{tab:app_campaign_hierarchy}. Solid bars show observed or clean endpoints, hatched segments unavailable endpoints; dark, medium, and light shading denote the primary resource panels, supporting campaigns, and the exploratory campaign. The strict-12k missing unit remains assigned in sharp-completion analyses. Historical clean exclusions are shown separately from valid outcomes. Both resource panels cover the same 22 mid-stratum and 18 ceiling-stratum tasks, with 240/240 endpoints observed in each workflow arm at each ceiling.
\end{minipage}
\end{figure}
\end{minipage}\par

\par\medskip
\noindent\begin{minipage}{\textwidth}
\subsection{Leave-One-Task-Out Stability}
\label{app:fig_loto}

Figure~\ref{fig:appendix_loto} reports leave-one-task-out estimates from the
planning-confirmation experiment.

\begin{figure}[H]
\singlespacing
\centering
\includegraphics[width=0.94\textwidth]
{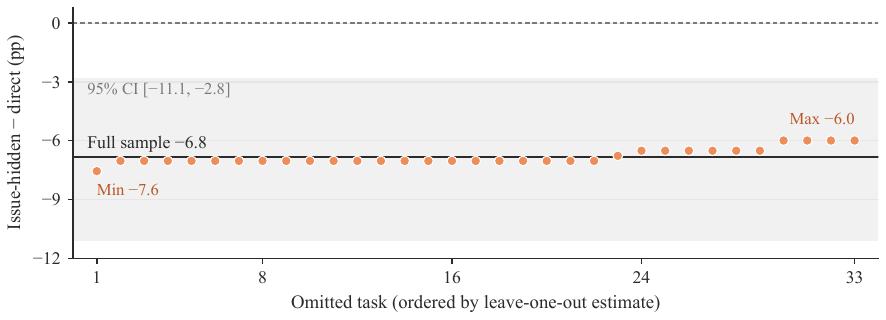}
\caption{Leave-One-Task-Out Stability}
\label{fig:appendix_loto}

\vspace{0.3em}
\begin{minipage}{0.94\textwidth}
\footnotesize
\textit{Notes:} Each point omits one of 33 task clusters and recomputes the equal-task mean $T_{GP}-T_G$ in the planning-confirmation sample. The full-sample line is $-6.8$ pp, and the shaded band is its 95\% task-cluster percentile-bootstrap interval, $[-11.1,-2.8]$ pp (2,000 resamples). Deletion estimates range from $-7.6$ to $-6.0$ pp. All deletions retain the negative sign and lie inside the full-sample interval; this is supporting stability evidence.
\end{minipage}
\end{figure}
\end{minipage}\par

\par\medskip
\noindent\begin{minipage}{\textwidth}
\subsection{Task-Level Planning Differences}
\label{app:fig_task_level}

Figure~\ref{fig:appendix_task_differences} displays task-level planning
contrasts at the two inference ceilings, linking each task across ceilings.

\begin{figure}[H]
\singlespacing
\centering
\includegraphics[width=0.94\textwidth]
{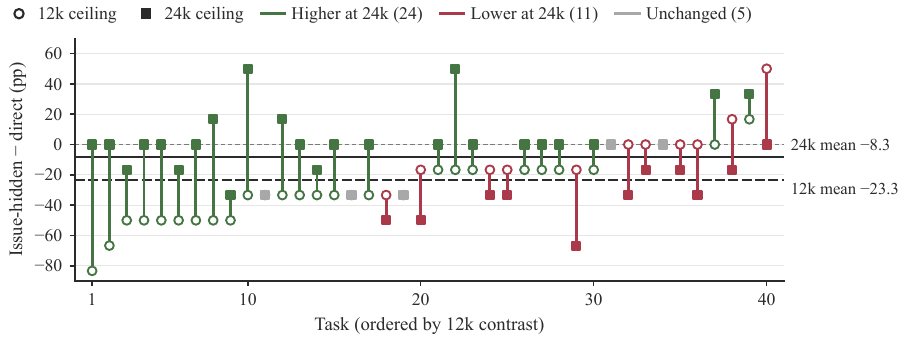}
\caption{Task-Level Planning Differences}
\label{fig:appendix_task_differences}

\vspace{0.3em}
\begin{minipage}{0.94\textwidth}
\footnotesize
\textit{Notes:} Each point is a task-level $T_{GP}-T_G$ contrast averaged over two backends and three replicates per arm. Each vertical segment connects one task's 12k contrast (open circle) and 24k contrast (filled square). Tasks are ordered by ascending 12k contrast, with ties broken by task identifier. Segment color shows whether the contrast is higher (24 tasks), lower (11), or unchanged (5) at 24k. Long-dashed and solid horizontal lines are the 12k and 24k means; the short-dashed line marks zero. The figure provides descriptive task-level dispersion.
\end{minipage}
\end{figure}
\end{minipage}\par

\par\medskip
\noindent\begin{minipage}{\textwidth}
\subsection{Protocol and Endpoint Sensitivity}
\label{app:fig_protocol_sensitivity}

Figure~\ref{fig:appendix_protocol_sensitivity} summarizes sensitivity to the
single missing information-treatment endpoint and displays the high-budget
task-informed campaign.

\begin{figure}[H]
\singlespacing
\centering
\includegraphics[width=0.94\textwidth]
{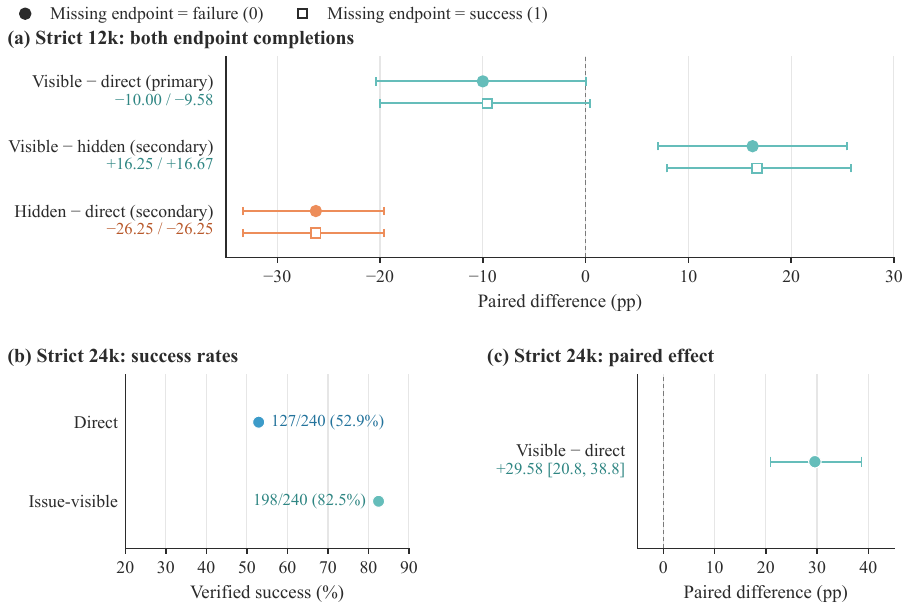}
\caption{Protocol and Endpoint Sensitivity}
\label{fig:appendix_protocol_sensitivity}

\vspace{0.3em}
\begin{minipage}{0.94\textwidth}
\footnotesize
\textit{Notes:} Panel (a) reports the three strict-12k protocol contrasts, retaining all assignments with the missing endpoint set to failure (filled circles) or success (open squares); bars are separate 95\% task-cluster percentile-bootstrap intervals (2,000 resamples). Panels (b) and (c) show the strict-24k campaign, with 240 observed endpoints per arm: panel (b) gives observed rates on a truncated axis, and panel (c) the paired $T_{GPA}-T_G$ contrast with its 95\% task-cluster percentile-bootstrap interval (2,000 resamples, same procedure as panel (a)).
\end{minipage}
\end{figure}
\end{minipage}\par

\par\medskip
\noindent\begin{minipage}{\textwidth}
\subsection{Descriptive Heterogeneity}
\label{app:fig_heterogeneity}

Figure~\ref{fig:appendix_heterogeneity} summarizes descriptive heterogeneity
across model backends and task-screening strata.

\begin{figure}[H]
\singlespacing
\centering
\includegraphics[width=0.94\textwidth]
{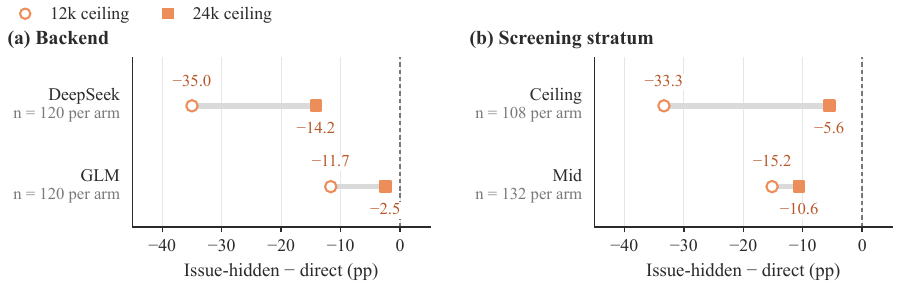}
\caption{Descriptive Heterogeneity}
\label{fig:appendix_heterogeneity}

\vspace{0.3em}
\begin{minipage}{0.94\textwidth}
\footnotesize
\textit{Notes:} Points show resource-panel $T_{GP}-T_G$ success contrasts by backend and screening stratum at 12k (open circles) and 24k (filled squares); segments connect the two ceilings within each subgroup, and row labels give observations per arm at each ceiling. DeepSeek and GLM refer to \texttt{deepseek-chat} and \texttt{glm-4.6}. Strata contain 18 ceiling and 22 mid tasks. These descriptive subgroup estimates do not establish statistically confirmed workflow-by-subgroup interactions.
\end{minipage}
\end{figure}
\end{minipage}\par

\end{document}